\documentclass[11pt,a4paper]{article}

\usepackage{authblk}
\usepackage[hyperref]{emnlp2020}
\usepackage{times}
\usepackage{latexsym}

\usepackage[utf8]{inputenc}
\usepackage{xcolor}
\usepackage{booktabs}

\usepackage{graphicx}

\usepackage{caption}
\usepackage{subcaption}

\usepackage{amssymb}

\usepackage{textcomp}

\usepackage{array}
\usepackage{rotating}
\newcolumntype{P}[2]{%
  >{\begin{turn}{#1}\begin{minipage}{#2}\small\raggedright}c%
  <{\end{minipage}\end{turn}}%
}

\usepackage{amsmath}

\DeclareMathOperator*{\entropy}{H}
\DeclareMathOperator{\expectation}{\mathbb{E}}
\DeclareMathOperator{\prob}{P}

\newcommand*{\Lbe}{\mathcal{L}_{\text{be}}}
\newcommand*{\Lie}{\mathcal{L}_{\text{ie}}}
\newcommand*{\x}{\mathbf{x}}

\newcommand*{\NPMI}{nPMI}
\newcommand*{\BCubed}{B\textsuperscript{3}}
\newcommand*{\diversity}{diversity}
\newcommand*{\Diversity}{Diversity}
\newcommand*{\DiversityShort}{Div}
\newcommand*{\uncertainty}{uncertainty}
\newcommand*{\Uncertainty}{Uncertainty}
\newcommand*{\UncertaintyShort}{Unc}

\usepackage{hyperref}
\usepackage{cleveref}

\usepackage{microtype}

\aclfinalcopy 
\def\aclpaperid{1052} 

\AddToShipoutPicture{%
  \AtPageLowishCenter{\textcolor{black}{\thepage}}
}

\newlength\stextwidth

\usepackage{todonotes}  
\usepackage[normalem]{ulem}  
\newif\ifcomments
\commentsfalse

\ifcomments
    \newcommand*{\TODO}[1]{\textcolor{red}{[TODO: #1]}}
    \newcommand*{\tocite}[1]{(\textcolor{blue}{#1})}
    \newcommand*{\tocitep}[1]{(\textcolor{blue}{#1})}
    \newcommand*{\tocitet}[1]{\textcolor{blue}{#1}}
    
    \newcommand*{\julian}[1]{\textcolor{orange}{[JJM: #1]}}
    \newcommand*{\ian}[1]{\textcolor{olive}{[IFT: #1]}}
    \newcommand*{\jan}[1]{\textcolor{purple}{[JAB: #1]}}
    
    \definecolor{iansnotecolor}{RGB}{255, 218, 181}
    \newcommand*{\iansidenote}[1]{
        \todo[color=iansnotecolor, size=\footnotesize]{%
        [\textbf{Ian:}] #1}%
    }

    \newcommand*{\maybedelete}[1]{\textcolor{red}{\sout{#1}}}
    
\else
    \newcommand*{\TODO}[1]{}
    \newcommand*{\tocite}[1]{}
    \newcommand*{\tocitep}[1]{}
    \newcommand*{\tocitet}[1]{}
    
    \newcommand*{\julian}[1]{}
    \newcommand*{\ian}[1]{}
    \newcommand*{\jan}[1]{}
    \newcommand*{\iansidenote}[1]{}
    
    \newcommand*{\maybedelete}[1]{}
    
\fi

\renewcommand\footnotemark{}
\makeatletter
\renewcommand\AB@authnote[1]{\rlap{\textsuperscript{\normalfont#1}}}

\makeatother

\title{Asking without Telling: Exploring Latent Ontologies \\ in Contextual Representations}

\author[1$*$]{Julian Michael\thanks{$^*$Work performed while at Google.}}
\author[2]{Jan A. Botha}
\author[2]{Ian Tenney}

\affil[1]{Paul G. Allen School of Computer Science \& Engineering, University of Washington}
\affil[2]{Google Research}

\affil[ ]{\texttt{julianjm@cs.washington.edu}}
\affil[ ]{\texttt{\{jabot,iftenney\}@google.com}}

\date{}

\begin{document}
\maketitle
\begin{abstract}

The success of pretrained contextual encoders, such as ELMo and BERT, has brought a great deal of interest in what these models learn: do they, without explicit supervision, learn to encode meaningful notions of linguistic structure?
If so, how is this structure encoded?
To investigate this, we introduce \textit{latent subclass learning} (LSL):
a modification to classifier-based probing that induces a latent categorization (or \textit{ontology}) of the probe's inputs.
Without access to fine-grained gold labels, LSL extracts \textit{emergent} structure from input representations in an interpretable and quantifiable form.
In experiments, we find strong evidence of familiar categories, such as a notion of personhood in ELMo, as well as novel ontological distinctions, such as a preference for fine-grained semantic roles on core arguments.
Our results provide unique new evidence of emergent structure in pretrained encoders, including departures from existing annotations which are inaccessible to earlier methods.

\end{abstract}

\section{Introduction}

The success of self-supervised pretrained models in NLP \cite{devlin2019bert,peters2018deep,radford2019language,lan2020albert} on many tasks \cite{wang2018glue, wang2019superglue}
has stimulated interest in how these models work, and what they learn about language.
Recent work on model analysis \citep{belinkov2019analysis}
indicates that they may learn a lot about linguistic structure, including
part of speech \citep{belinkov2017neural},
syntax \citep{blevins2018hierarchical,marvin2018targeted},
word sense \citep{peters2018deep,reif2019bertviz},
and more \citep{rogers2020primer}.

\begin{figure}[t!]
  \includegraphics[width=\columnwidth]{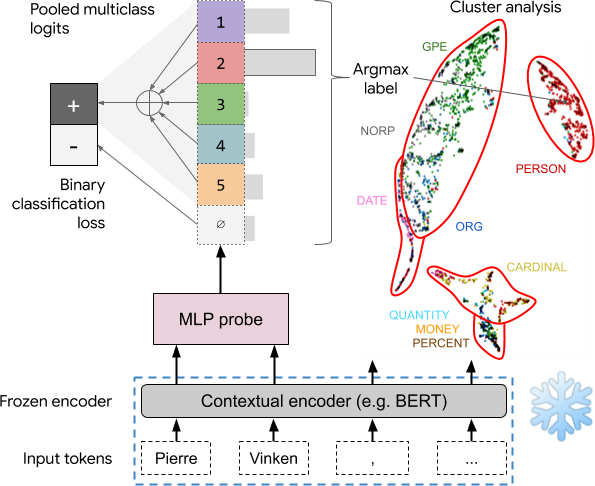}
  \caption{LSL overview. A probing classifier over contextual embeddings produces multi-class \textit{latent logits}, which are marginalized into a single logit trained on binary classification. In this example, ``Pierre Vinken'' is identified as a named entity and assigned to latent class 2, which aligns well with the PERSON label. We treat the classes as clusters representing a latent ontology that describes the underlying representation space.
  \autoref{fig:ner-vis} visualizes latent logits in more detail.}
  \label{fig:schematic}
  \vspace{-1em}
\end{figure}

Many of these results are based on \textit{predictive methods}, such as probing, which measure how well a linguistic variable can be predicted from intermediate representations.
However,
the ability of supervised probes to fit weak features makes it difficult to produce unbiased answers
about how those representations are structured~\citep{saphra-lopez-2019-understanding,voita2019bottom}.
\textit{Descriptive methods} like clustering and visualization explore this structure directly, but provide limited control and often regress to dominant categories such as lexical features \citep{singh2019bert} or word sense \citep{reif2019bertviz}. This leaves open many questions: \textit{how} are linguistic features like entity types, syntactic dependencies, or semantic roles represented by an encoder like ELMo \citep{peters2018deep} or BERT \citep{devlin2019bert}?
To what extent do familiar categories like PropBank roles or Universal Dependencies appear naturally?
Do these unsupervised encoders learn their own categorization of language?

To tackle these questions, we propose a
systematic way to extract \textit{latent ontologies}, or discrete categorizations of a representation space, which we call \textit{latent subclass learning} (LSL); see \autoref{fig:schematic} for an overview.
In LSL, we use a binary classification task (such as detecting entity mentions or syntactic dependency arcs) as weak supervision to induce a set of latent clusters relevant to that task (\textit{i.e.}, entity or dependency types).
As with \textit{predictive} methods, the choice of task allows us to explore varied phenomena, and induced clusters can be quantified and compared to gold annotations. But also, as with \textit{descriptive} methods, our clusters can be inspected and qualified directly, and observations have high specificity: agreement with external (\textit{e.g.}, gold) categories provides strong evidence that those categories are salient in the representation space.

We describe the LSL classifier in \autoref{sec:approach}, and apply it to the edge probing paradigm \citep{tenney2019what} in \autoref{sec:experimental-setup}.
In \autoref{sec:results} we evaluate LSL on multiple encoders, including ELMo
and BERT.
We find that LSL induces stable and consistent ontologies, which include both striking rediscoveries of gold categories---for example, ELMo discovers \textit{personhood} of named entities and BERT has a notion of \textit{dates}---and novel ontological distinctions---such as fine-grained core argument semantic roles---which are not easily observed by fully supervised probes.
Overall, we find unique new evidence of emergent latent structure in our encoders, while also revealing new properties of their representations which are inaccessible to earlier methods.

\begin{figure*}
  \centering
      \centering
      \includegraphics[width=.9\textwidth]{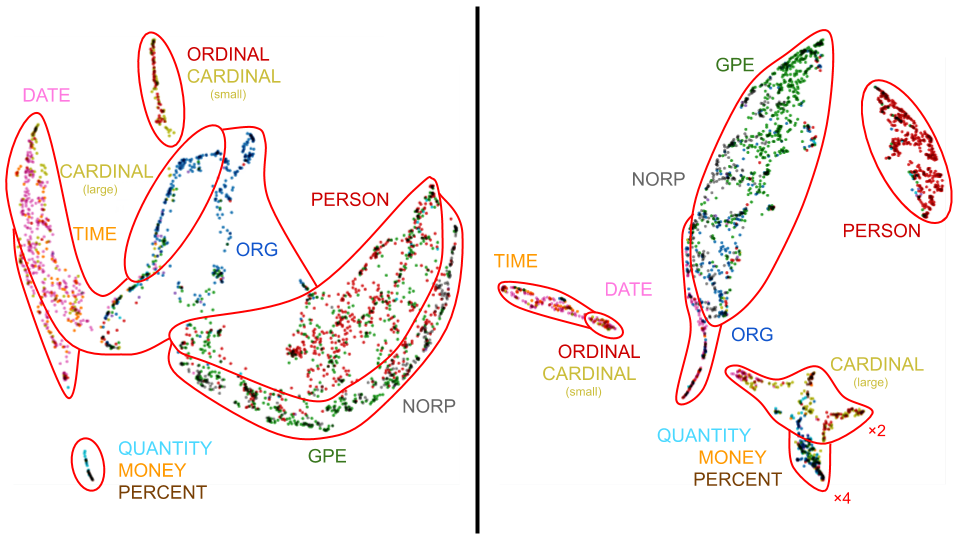}
      \caption{
      Latent logit vectors from BERT (left) and ELMo (right)
      for a sample from the Named Entities development set
      visualized in the Embedding Projector \citep{smilkov2016embedding}
      using UMAP \citep{mcinnes2018umap}, which is designed to preserve local clustering structure in a low dimensional visualization.
      Points are colored by gold label, and induced clusters are outlined in red.
      ELMo has a clear notion of personhood (PERSON),
      while BERT groups people with geopolitical entities (GPE) and nationalities (NORP).
      BERT strongly identifies dates (DATE) and organizations (ORG),
      and both models group numeric/quantitative entities together.
      Both models separate small CARDINAL numbers (roughly, seven or less) and group them with ORDINALs,
      separate from larger CARDINALs.
      The outlined areas in the bottom-right of the ELMo visualization include 2 and 4 induced clusters. }
      \label{fig:ner-vis}
\end{figure*}

\section{Background}
\label{sec:background}


\paragraph{Predictive analysis}

A common form of model analysis is \textit{predictive}:
assessing how well a linguistic variable can be predicted from a model,
whether in intrinsic behavioral tests \citep{goldberg2019assessing,marvin2018targeted,petroni2019language} or extrinsic \textit{probing tasks}.

\textit{Probing} involves training lightweight classifiers over features produced by a pretrained model, and
assessing the model's knowledge by the probe's performance.
Probing has been used for low-level properties such as word order and sentence length \citep{adi2016fine,conneau2018cram},
as well as phenomena at the level of syntax \citep{hewitt-manning-2019-structural},
semantics \citep{tenney2019what,liu2019multi,clark2019what},
and discourse structure \citep{chen-etal-2019-evaluation}.
Error analysis on probes has been used to argue
that BERT may simulate sequential decision making across layers \citep{tenney2019bert},
or that it encodes its own, soft notion of syntactic distance \citep{reif2019bertviz}.

Predictive methods such as probing are \textit{flexible}:
Any task with data can be assessed.
However, they only track predictability of pre-defined categories, limiting their descriptive power.
In addition, a powerful enough probe, given enough data,
may be insensitive to differences between encoders,
making it difficult to interpret results based on
accuracy \citep{saphra-lopez-2019-understanding,zhang2018}.
So, many probing experiments appeal to the \textit{ease of extraction} of a linguistic variable \citep{pimentel2020informationtheoretic}.
Existing work has measured this by controlling for probing model capacity,
either using relative claims between layers and encoders
\citep{belinkov2017evaluating,blevins2018hierarchical,tenney2019what,liu2019linguistic}
or using explicit measures to estimate and trade off capacity with accuracy
\citep{hewitt-liang-2019-designing,voita2020informationtheoretic}.
An alternative is to control
\textit{amount of supervision}, by
restricting training set size \citep{zhang2018},
comparing learning curves \citep{talmor2019olmpics},
or using description length with online coding \citep{voita2020informationtheoretic}.

We extend this further by removing the distinction between gold categories in the training data and reducing the supervision to binary classification, as explained in \autoref{sec:approach}.
This extreme measure makes our test \textit{high specificity}, in the sense that positive results---\textit{i.e.}, when comprehensible categories are recovered by our probe---are much stronger, since a category must be essentially \textit{invented} without direct supervision.

\paragraph{Descriptive analysis}

In contrast to predictive methods, which assess an encoder with respect to particular data, \textit{descriptive} methods analyze models on their own terms,
and include clustering, visualization \citep{reif2019bertviz}, and correlation analysis techniques \citep{voita2019bottom,saphra-lopez-2019-understanding,abnar2019blackbox,chrupala2019correlating}.
Descriptive methods produce high-specificity tests of what structure is present in the model, and facilitate discovery of new patterns that were not hypothesized prior to testing.
However, they lack the flexibility of predictive methods.
Clustering results tend to be dominated by principal components of the embedding space, which correspond to only some salient aspects of linguistic knowledge,
such as lexical features \citep{singh2019bert} and word sense \citep{reif2019bertviz}.
Alternatively, more targeted analysis techniques generally have a restricted inventory of inputs, such as layer mixing weights \citep{peters2018dissecting}, transformer attention distributions \citep{clark2019what}, or pairwise influence between tokens \citep{wu2020perturbed}.
As a result of these issues, it is more difficult to discover the underlying structure corresponding to rich, layered ontologies.
Our approach retains the advantages of descriptive methods,
while admitting more control as the choice of binary classification targets can
guide the LSL model to discover structure relevant to a particular linguistic task.



\paragraph{Linguistic ontologies}
Questions of what encoders learn about language require well-defined \textit{linguistic ontologies},
or meaningful categorizations of inputs, to evaluate against.
Most analysis work uses formalisms from the classical NLP pipeline,
such as part-of-speech and syntax from the Penn Treebank \citep{marcus1993building}
or Universal Dependencies \citep{nivre2015universal},
semantic roles from PropBank \citep{palmer2005proposition}
or \citet{dowty1991thematic}'s Proto-Roles \citep{reisinger2015semantic},
and named entities, which have a variety of available ontologies
\citep{pradhan2007ontonotes,ling2012finegrained,choi2018ultra}.
Work on ontology-free, or \textit{open}, representations suggests that the linguistic structure
captured by traditional ontologies may be encoded in a variety of possible ways
\citep{banko2007open,he2015question,michael2018crowdsourcing}
while being annotatable at large scale \citep{fitzgerald2018large}.
This raises the question:
when looking for linguistic knowledge in pretrained encoders, what exactly should we expect to find?
Predictive methods are useful for fitting an encoder to an existing ontology;
but do our encoders latently hold their own ontologies as well?
If so, what do they look like?
That is the question we investigate in this work.

\section{Approach}
\label{sec:approach}

We propose a way to extract latent linguistic ontologies from pretrained encoders
and systematically compare them to existing gold ontologies.
We use a classifier based on \textit{latent subclass learning} (\autoref{sec:latent-subclass-learning}),
which is applicable in any binary classification
setting.\footnote{A similar classifier was concurrently
developed and presented for use in model distillation by
\citet{muller2020subclass}.}
We propose several quantitative metrics to evaluate the induced ontologies (\autoref{sec:metrics}),
providing a starting point for qualitative analysis (\autoref{sec:results}) and future research.

\subsection{Latent Subclass Learning}
\label{sec:latent-subclass-learning}

Consider a logistic regression classifier over inputs \(\mathbf{x} \in \mathbb{R}^d \).
It outputs probabilities according to the following formula:
\[
\prob(y \mid \mathbf{x}) = \sigma(\mathbf{w}^\top \mathbf{x}),
\]
where \( \mathbf{w} \in \mathbb{R}^d \) is a learned parameter.
Instead, we propose the \textit{latent subclass learning} classifier:
\[
\prob_{\mathrm{LSL}}(y \mid \mathbf{x}) = \sigma\left(\log \sum_i^N e^{\mathbf{W}_i\mathbf{x}} \right),
\]
where \( \mathbf{W} \in \mathbb{R}^{N \times d} \) is a parameter matrix,
and \( N \) is a hyperparameter corresponding to the number of latent classes.

This corresponds to $N$+1-way multiclass logistic regression with a fixed 0 baseline for a null class,
but trained on binary classification by marginalizing over the $N$ non-null classes (\autoref{fig:schematic}).
The vector \(\mathbf{Wx} \in \mathbb{R}^N \) may then be treated as a set of \textit{latent logits}
for a random variable \(C(\mathbf{x}) \in \{1, \dots, N \}\) defined by the softmax distribution.
Taking the hard maximum of \(\mathbf{Wx}\) assigns a latent class \(\hat{C}(\x)\) to each input,
which may be viewed as a \textit{weakly supervised clustering},
learned on the basis of external supervision
but not explicitly optimized to match prior gold categories.

For the loss $\mathcal{L}_\textrm{LSL}$, we use the cross-entropy loss on $\prob_{\mathrm{LSL}}$.
However, this does not necessarily encourage a diverse, coherent set of clusters;
an LSL classifier may simply choose to collapse all examples into a single category,
producing an uninteresting ontology.
To mitigate this, we propose two \textit{clustering regularizers}.

\paragraph{Adjusted batch-level negative entropy}
We wish for the model to induce a diverse ontology.
One way to express this is that the expectation of \(C\) has high entropy,
\textit{i.e.}, we wish to maximize
\[
\entropy(\expectation_\mathbf{x} C(\mathbf{x})).
\]
In practice, we use the expectation over a batch.
The maximum value this can take is the entropy of the uniform distribution over \( N \) items,
or \( \log N \).
Therefore, we wish to minimize the \textit{adjusted batch-level negative entropy loss}:
\[ \Lbe = \log N - \entropy(\expectation_\mathbf{x} C(\mathbf{x})),  \]
which takes values in \([0, \log N]\).

\paragraph{Instance-level entropy}
In addition to using all latent classes in the expected case, we also wish for the model to assign
a single coherent class label to each input example.
This can be done by minimizing the \textit{instance-level entropy loss}:
\[ \Lie = \expectation_\mathbf{x} \entropy(C(\mathbf{x})). \]
This also takes values in \([0, \log N]\),
and we compute the expectation over a batch.

\paragraph{Loss}
We optimize the regularized LSL loss
\[
\mathcal{L}_{\mathrm{LSL}} + \alpha \Lbe + \beta \Lie,
\]
where $\alpha$ and $\beta$ are hyperparameters, via gradient descent.
Together, the regularizers encourage a balanced solution
where the model uses many clusters yet gives each input a distinct assignment.
Note that if $\alpha = \beta$, the this objective maximizes the mutual information between $\mathbf{x}$ and $C$, encouraging the ontology to encode as much information as possible about the training data while still supporting the binary classification objective.

\subsection{Metrics}
\label{sec:metrics}

Since our interest is in descriptively analyzing encoders' latent ontologies, there are no normatively `correct' categories. However, we can leverage existing gold ontologies---such as PropBank role labels or Universal Dependencies---to quantify our results in terms of well-understood categories.
For the following metrics, we consider only points in the gold positive class.

\paragraph{\BCubed{}}
B-cubed (or \BCubed{}) is a standard clustering metric \citep{bagga-baldwin-1998-bcubed,amigo2009comparison} which calculates the precision and recall of each point's predicted cluster against its gold cluster, averaging over points.
It allows for label-wise scoring by restricting to points with specific gold labels, allowing for fine-grained analysis, \textit{e.g.}, of whether a gold label is concentrated in few predicted clusters (high recall) or well-separated from other labels (high precision).

\paragraph{Normalized PMI}
Pointwise mutual information (PMI) is commonly used as an association measure
reflecting how likely two items (such as tokens in a corpus) are to occur together
relative to chance \citep{church1990word}.
\textit{Normalized} PMI  \citep[\NPMI{};][]{bouma2009normalized}
is a way of factoring out the effect of item frequency on PMI.
Formally, the \NPMI{} of two items \(x\) and \(y\) is
\[
\left. \left(\log \frac{\prob(x,y)}{\prob(x)\prob(y)} \right) \middle/ -\log(\prob(x,y)) \right.,
\]
taking the limit value of -1 when they never occur together,
1 when they only occur together, and 0 when they occur independently.
We use \NPMI{} to analyze the co-occurrence of \textit{gold labels} in \textit{predicted clusters}:
A pair of gold labels with high \NPMI{} are preferentially grouped together by the induced ontology,
whereas two labels with low \NPMI{} are preferentially distinguished.

Plotting pairwise \NPMI{} of gold labels allows us to see specific ways the induced clustering agrees or disagrees with a gold reference (\autoref{sec:results}, \autoref{fig:npmi-megafigure}).
Since \NPMI{} is information-theoretic and chance-corrected, it is a reliable indicator of the degree of information about gold labels contained in a set of predicted clusters. However, it is relatively insensitive to cluster granularity (\textit{e.g.}, the total number of predicted categories, or whether a single gold category is split into many different predicted clusters), which is better understood through our other metrics.


\paragraph{\Diversity}
We desire fine-grained ontologies with many meaningful classes.
Number of attested classes may not be a good measure of this,
since it could include classes with very few members and no broad meaning.
So we propose \textit{\diversity}:
\[
\exp(\entropy(\expectation_\mathbf{x} \hat{C}(\mathbf{x}))).
\]
This increases as the clustering becomes more fine-grained and evenly distributed,
with a maximum of \(N\) when \(\prob(\hat{C})\) is uniform.
More generally, exponentiated entropy is sometimes referred to as the \textit{perplexity} of a distribution,
and corresponds (softly) to the number of classes required for a uniform distribution of the same entropy.
In that sense, it may be regarded as the effective number of classes in an ontology.
We use the predicted class \(\hat{C}\) rather than its distribution \(C\) because we care about
the diversity of the model's clustering, and not just uncertainty in the model.

\paragraph{\Uncertainty}
In order for our learned classes to be meaningful, we desire distinct and coherent clusters.
To measure this, we propose \textit{\uncertainty}:
\[
\expectation_{\mathbf{x}} \exp(\entropy(C(\mathbf{x}))).
\]
This is also related to perplexity, but unlike \diversity{},
it takes the expectation over the input after calculating the perplexity of the distribution.
This reflects how many classes, on average, the model is confused between when provided with an input.
Low values correspond to coherent clusters,
with a minimum of 1 when every latent class is assigned with full confidence.
As with \diversity{}, we take the expectation over the evaluation set.

\begin{table*}
\centering
\begin{tabular}{lrrrr|rrrrr}
\toprule
& \multicolumn{4}{c|}{Named Entities}
& \multicolumn{4}{c}{Universal Dependencies} \\
\midrule
                    & \textbf{P / R / F1}       & \textbf{Acc.} &  \textbf{\DiversityShort$\uparrow$} & \textbf{\UncertaintyShort$\downarrow$}
                    & \textbf{P / R / F1}       & \textbf{Acc.} & \textbf{\DiversityShort$\uparrow$} & \textbf{\UncertaintyShort$\downarrow$} \\

\midrule
\textbf{Gold}       & 1.0 / 1.0 / 1.0           & 1.0 & 9.71 & 1.00
                    & 1.0 / 1.0 / 1.0           & 1.0 & 22.91 & 1.00 \\
\midrule
\textbf{Multi}      & .86 / .88 / .87           & .94 & 8.58 & 1.88 
                    & .86 / .83 / .84           & .93 & 21.94 & 1.77 \\
\midrule
\textbf{LSL}        & .28 / .80 / .41           & .96 & 2.85 & 1.45 
                    & .10 / .60 / .18           & .94 & 3.50 & 2.07 \\
\textbf{\, +be}     & .20 / .43 / .27           & .96 & 4.78 & 31.23 
                    & .18 / .13 / .15           & .94 & \textbf{29.83} & 12.33 \\
\textbf{\, +ie}     & .13 / 1.0 / .23           & .93 & 1.00 & \textbf{1.00}
                    & .09 / .79 / .15           & .94 & 2.00 & \textbf{1.01} \\
\textbf{\, +be +ie}  & .43 / .54 / \textbf{.48}  & .88 &\textbf{7.00} & 1.10
                    & .18 / .27 / \textbf{.22}  & .86 & 14.96 & 1.35 \\
\midrule
\textbf{Single}     & .13 / 1.0 / .23 & - & 1.00 & 1.00
                    & .06 / 1.0 / .11 & - & 1.00 & 1.00 \\
\bottomrule
\end{tabular}
\caption{
    Model selection results over BERT-large.
    \textbf{Multi} is the standard multi-class model trained directly on gold labels, and
    \textbf{Single} is the degenerate single-cluster baseline.
    Our clustering regularizers (\textbf{b}atch and/or \textbf{i}nstance-level \textbf{e}ntropy), when taken together, yield a good tradeoff between \textbf{div}ersity and \textbf{unc}ertainty,
    though at some expense to binary classification \textbf{acc}uracy. 
}
\label{tab:model-selection-results}
\end{table*}

\section{Experimental Setup}
\label{sec:experimental-setup}

We adopt a similar setup to \citet{tenney2019what} and \citet{liu2019linguistic},
training probing models over several contextualizing encoders on a variety of linguistic tasks.

\subsection{Tasks}

We cast several structure labeling tasks from \citet{tenney2019what} as binary classification
by adding negative examples, bringing the positive to negative ratio to 1:1 where possible.

\vspace{0.5em}\noindent
\textbf{Named entity labeling} requires labeling
noun phrases with
entity types, such as person, location, date, or time.
We randomly sample non-entity noun phrases as negatives.

\vspace{0.5em}\noindent
\textbf{Nonterminal labeling} requires labeling
phrase structure constituents with 
syntactic types, such as noun phrases and verb phrases.
We randomly sample non-constituent spans as negatives.

\vspace{0.5em}\noindent
\textbf{Syntactic dependency labeling} requires labeling
token pairs with their syntactic relationship,
such as a subject, direct object, or modifier.
We randomly sample non-attached token pairs as negatives.

\vspace{0.5em}\noindent
\textbf{Semantic role labeling} requires labeling
predicates (usually verbs) and their arguments (usually syntactic constituents)
with labels that abstract over syntactic relationships
in favor of more semantic notions such as \textit{agent}, \textit{patient},
modifier roles involving, \textit{e.g.}, time and place, or predicate-specific roles.
We draw the closest non-attached predicate-argument pairs as negatives.

\vspace{0.5em}\noindent
We use the English Web Treebank part of Universal Dependencies 2.2 \citep{silveira14gold}
for dependencies, and the English portion of Ontonotes~5.0 \citep{weischedel2013ontonotes} for other tasks.

\subsection{Encoders}

We run experiments on the following encoders:

\vspace{0.5em}\noindent
\textbf{ELMo}
\citep{peters2018deep} is the concatenation of representations from 2-layer LSTMs  \citep{hochreiter1997long} trained with forward and backward language modeling objectives.
We use the publicly available instance\footnote{\url{tfhub.dev/google/elmo/2}} trained on the One Billion Word Benchmark \citep{chelba2014one}.

\vspace{0.5em}\noindent
\textbf{BERT}
\citep{devlin2019bert} is a deep Transformer stack \cite{vaswani2017attention}
trained on masked language modeling and next sentence prediction tasks.
We use the 24-layer BERT-large instance\footnote{
\url{github.com/google-research/bert} 
}
trained on about 2.3B tokens from
English Wikipedia and BooksCorpus \citep{zhu2015aligning}.

\vspace{0.5em}\noindent
\textbf{BERT-lex} is a lexical baseline, using only BERT's context-independent wordpiece embedding layer.

\subsection{Probing Model}

We reimplement the model of \citet{tenney2019what},\footnote{Publicly available at \url{https://jiant.info}} which gives a unified architecture that works for a wide range of probing tasks. Specifically, it classifies single spans or pairs of spans in the following way:
1) construct token representations by pooling across encoder layers with a learned scalar mix \citep{peters2018deep},
2) construct span representations from these token representations using self-attentive pooling \citep{lee2017end}, and
3) concatenate those span representations and feed the result through a fully-connected layer to produce input features for the classification layer.
We follow \citet{tenney2019what} in training a probing model over a frozen encoder, while using our LSL classifier (\autoref{sec:approach}) as the final output layer in place of the usual softmax.

\subsection{Model selection}
\label{sec:model-selection}
We run initial studies to determine hidden layer sizes and regularization coefficients.
For all LSL probes, we use \(N = 32\) latent classes.\footnote{Preliminary experiments
found similar results for larger \(N\), with similar diversity in the full setting.}

\paragraph{Probe capacity}
To mitigate the influence of probe capacity on the results, we follow the best practice recommended by \citet{hewitt-liang-2019-designing} and use a single hidden layer with the smallest size that does not sacrifice performance.
For each task, we train binary logistic regression probes with a range of hidden sizes
and select the smallest yielding at least 97\% of the best model's performance.
Details are in Appendix A.

\paragraph{Mitigating variance}
To decrease variance across random restarts,
we use a consistency-based model selection criterion:
train 5 models, compute their pairwise \BCubed{} F1 scores, and choose the one with the highest average F1.
(However, as we find in \autoref{sec:results}, the qualitative patterns that emerged were consistent between runs.)

\paragraph{Regularization coefficients}
We run preliminary experiments using BERT-large on Universal Dependencies and Named Entity Labeling
with ablations on our clustering regularizers.
For each ablation, we choose hyperparameters with the best F1 against gold.

\begin{table*}
\centering
\begin{tabular}{l|rr|rr|rr|r}

\toprule
     & \multicolumn{2}{|c|}{\textbf{BERT-lex}}
     & \multicolumn{2}{|c|}{\textbf{ELMo}}
     & \multicolumn{2}{|c|}{\textbf{BERT-large}}
     & \multicolumn{1}{|c }{\textbf{Gold}}
     \\
\midrule
\textbf{Task}
    & P / R / F1
    & \DiversityShort
    & P / R / F1
    & \DiversityShort
    & P / R / F1
    & \DiversityShort
    & \DiversityShort
    \\
\midrule

\textbf{Dependencies}
    & .06 / .86 / .11
    & 1.33
    & .23 / .42 / \textbf{.29}
    & 11.11
    & .14 / .33 / .19
    & \textbf{11.22}
    & 22.91
    \\
\textbf{Named Entities}
    & .19 / .39 / .26
    & 4.33
    & .40 / .66 / \textbf{.50}
    & 5.07
    & .47 / .53 / \textbf{.50}
    & \textbf{7.50}
    & 9.71
    \\
\textbf{Nonterminals}     
    & .22 / .80 / .34
    & 1.47
    & .36 / .25 / .30
    & \textbf{10.16}
    & .35 / .34 / \textbf{.35}
    & 7.80
    & 7.15
    \\
\textbf{Semantic Roles}   
    & .19 / .39 / \textbf{.26}
    & 2.81
    & .40 / .17 / .24
    & \textbf{22.35}
    & .37 / .17 / .24 
    & 18.70 
    & 8.73 
    \\

\bottomrule
\end{tabular}
\caption{Results by task for three pretrained encoding methods.
All probing models were trained with the LSL loss and cluster regularization coefficients \( \alpha = \beta = 1.5 \),
and chosen by the best-of-5 consistency criterion and detailed in \autoref{sec:model-selection}.
Uncertainty for all models was close to 1 and is omitted for space.}
\label{tab:full-experiment-results}
\end{table*}

\paragraph{Results}
Results are shown in \autoref{tab:model-selection-results}.
As expected, the batch-level entropy loss drives up both diversity and uncertainty,
while the instance-level entropy loss drives them down.
In combination, however, they produce the right balance,
with uncertainty near 1 while retaining diversity.

Notably, the Named Entity model with the batch-level loss has \textit{higher} diversity when the instance-level loss is added.
This happens because batch-level entropy
can be increased by driving up instance-level entropy
without changing the entropy of the expected distribution of predictions
\(\entropy(\expectation_\x \prob(\hat{C}(\x)))\).
So by keeping the uncertainty down on each input,
the instance-level entropy loss helps the batch-level entropy loss
promote diversity in the induced ontology.

Based on these results, we set $\alpha = \beta = 1.5$ for $\mathcal{L}_{be}$ and $\mathcal{L}_{ie}$ for the main experiments.

\section{Results and Analysis}
\label{sec:results}

\autoref{tab:full-experiment-results} shows aggregate results for the tasks and encoders described in
\autoref{sec:experimental-setup}.\footnote{Results for more tasks and encoders are in Appendix B.}
Taking all metrics into account, contextualized encodings
produce richer ontologies that agree more with gold than the lexical baseline does.
In fact, BERT-lex has normalized PMI scores very close to zero across the board (plots are provided in Appendix C), encoding virtually no information about gold categories.
For this reason, we omit it from the rest of the analysis.




\paragraph{Named entities}
As shown in \autoref{tab:ner-f1-results},
neither BERT nor ELMo are sensitive to categories that are related to specialized world knowledge,
such as languages, laws, and events.
However, they are in tune with other types:
ELMo discovers a clear PERSON category,
whereas BERT has distinguished DATEs.
Visualization of the clusters (\autoref{fig:ner-vis}) corroborates this,
furthermore showing that the models have a sense of scalar values and measurement;
indeed, instead of the gold distinction between ORDINAL and CARDINAL numbers,
both models distinguish between \textit{small} and \textit{large} (roughly, seven or greater) numbers.
See Appendix C for detailed \NPMI{} scores.

\begin{figure}
  \centering
  \begin{subfigure}[b]{\columnwidth}
      \centering
      \includegraphics[height=0.245\textheight]{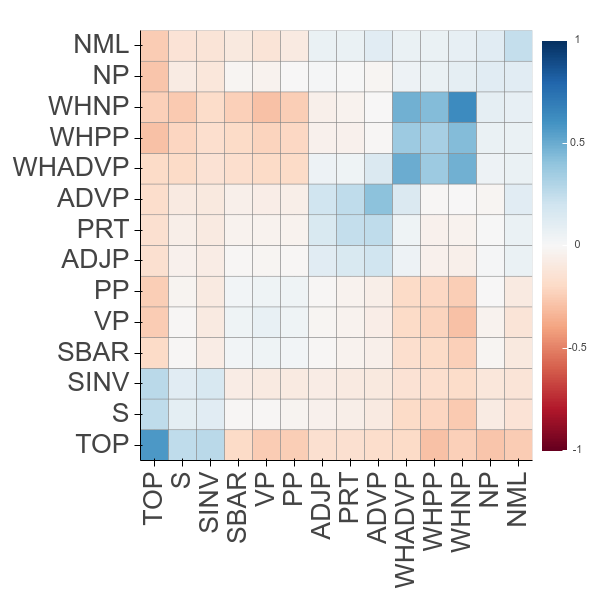}
      \caption{Nonterminals.}
      \label{fig:nonterminal-npmi-cmp}
  \end{subfigure}
  \begin{subfigure}[b]{\columnwidth}
      \centering
      \includegraphics[height=0.245\textheight]{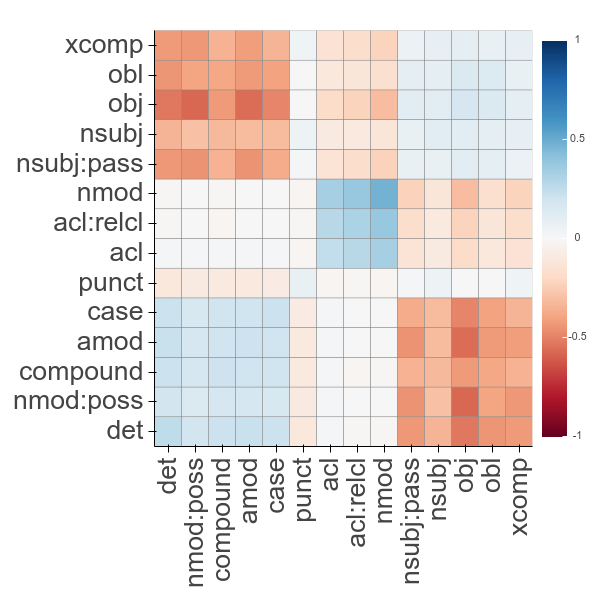}
      \caption{Universal dependencies.}
      \label{fig:ud-npmi-cmp}
  \end{subfigure}
  \begin{subfigure}[b]{\columnwidth}
      \centering
      \includegraphics[height=0.245\textheight]{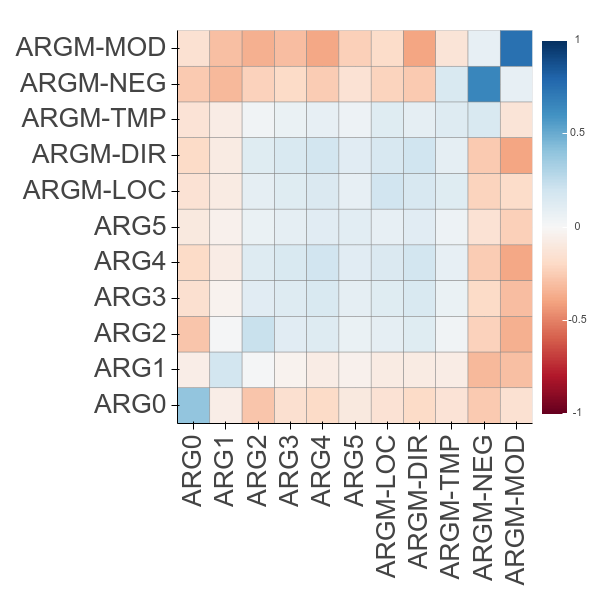}
      \caption{Semantic roles.}
      \label{fig:srl-npmi-cmp}
  \end{subfigure}
  \caption{Pairwise gold label \NPMI{}s on selected categories for ontologies induced from BERT-large on selected tasks.
  Blue is positive \NPMI{}, representing that gold labels are preferentially grouped together (\textit{i.e.}, conflated by the model) relative to chance.
  Red is negative \NPMI{}, representing that gold labels are well-separated.
  Perfectly matching ontologies would be 1 (blue) along the diagonal and -1 (red) in all off-diagonal cells.
  Counts are summed over all 5 runs to better reflect the underlying representations,
  though variance was low and our observed trends hold across all runs.
  }
  \label{fig:npmi-megafigure}
\end{figure}

\paragraph{Nonterminals}
Patterns in \NPMI{} (\autoref{fig:nonterminal-npmi-cmp})
suggest basic syntactic notions:
complete clauses (S, TOP, SINV) form a group,
as do phrase types which take subjects (SBAR, VP, PP),
and wh-phrases (WHADVP, WHPP, WHNP).

\paragraph{Dependencies}
Patterns in \NPMI{} (\autoref{fig:ud-npmi-cmp}) indicate several salient groups:
verb arguments (nsubj, obj, obl, xcomp),
left-heads (det, nmod:poss, compound, amod, case),
right-heads (acl, acl:relcl, nmod\footnote{Often the object in a prepositional phrase modifying a noun.}),
and punct.

\addtolength{\tabcolsep}{-1pt}
\begin{table}
\centering
\begin{tabular}{lcccccccc}
\toprule
 &
    \multicolumn{1}{P{90}{8ex}}{DATE} &
    \multicolumn{1}{P{90}{8ex}}{PERCENT} &
    \multicolumn{1}{P{90}{8ex}}{ORG} &
    \multicolumn{1}{P{90}{8ex}}{PERSON} &
    \multicolumn{1}{P{90}{8ex}}{\dots} &
    \multicolumn{1}{P{90}{8ex}}{EVENT} &
    \multicolumn{1}{P{90}{8ex}}{LAW} &
    \multicolumn{1}{P{90}{8ex}}{LANG.} \\
\midrule
BERT &
  .70 & 
  .60 & 
  .54 &
  .48 &
  $\cdot$ &
  .03 &
  .02 &
  .01  \\
ELMo &
  .38 &
  .28 &
  .35 &
  .81 & 
  $\cdot$ &
  .02 &
  .01 &
  .01 \\
\bottomrule
\end{tabular}
\caption{Label-wise \BCubed{} F1 scores for Named Entities, sorted by decreasing BERT-large F1.
Induced ontologies capture some labels surprisingly well,
but are indifferent to more specialized categories which may require more world knowledge to distinguish.
}
\label{tab:ner-f1-results}
\vspace{-0.25em}
\end{table}
\addtolength{\tabcolsep}{1pt}

\paragraph{Semantic roles}
Patterns in \NPMI{} (\autoref{fig:srl-npmi-cmp}) roughly match intuition:
primary core arguments (ARG0, ARG1) are distinguished,
as well as modals (ARGM-MOD) and negation (ARGM-NEG),
while trailing arguments (ARG2--5) and modifiers (ARGM-TMP, LOC, etc.) form a large group.
On one hand, this reflects surface patterns:
primary core arguments are usually close to the verb, with ARG0 on the left and ARG1 on the right;
trailing arguments and modifiers tend to be prepositional phrases or subordinate clauses;
and modals and negation are identified by lexical and positional cues.
On the other hand, this also reflects error patterns in state-of-the-art systems,
where label errors can sometimes be traced to ontological choices in
PropBank, which distinguish between arguments and adjuncts that have very similar meaning
\citep{he2017deep,kingsbury2002adding}.

While the number of induced classes roughly matches gold for most tasks,
induced ontologies for semantic roles are considerably more diverse, with a diversity measure close to 20 for ELMo and BERT (\autoref{tab:full-experiment-results}).
Even though the alignment of predicted clusters with gold is dominated by a few patterns (\autoref{fig:npmi-megafigure}), the induced clustering contains more information than just these patterns.
To locate this information, we examine the gold classes exhibiting the highest B$^3$ precision, shown in \autoref{tab:srl-split-categories}. Among these, core arguments ARG0, ARG1, and ARG2 have very low recall, indicating that the ontology splits them into finer-grained labels.

This follows intuition for PropBank core argument labels,
which have predicate-specific meanings.
Other approaches based on 
Frame Semantics \citep{baker1998berkeley,fillmore2006frame},
Proto-Roles \citep{dowty1991thematic,reisinger2015semantic},
or Levin classes \citep{levin1993english,schuler2005verbnet}
have more explicit fine-grained roles.
Concurrent work \citep{kuznetsov2020matter} shows that the choice of semantic role formalism meaningfully affects the behavior of supervised probes; further comparisons using LSL probing may help shed light on the origins of such differences. 

\begin{table}
\centering
\begin{tabular}{lr}
\toprule
Gold Label & P / R / F1 \\
\midrule
ARGM-MOD & .62 / .41 / .49 \\
ARG0     & .52 / .17 / .26 \\
ARG1     & .50 / .09 / .15 \\
ARGM-NEG & .36 / .60 / .45 \\
ARG2     & .28 / .13 / .18 \\
\bottomrule
\end{tabular}
\caption{Top semantic role labels by BERT-large \BCubed{} precision.
Core arguments ARG0--2 are most preferentially split, with high precision but low recall.}
\label{tab:srl-split-categories}
\vspace{-0.25em}
\end{table}

\section{Discussion}
\label{sec:discussion}

Our exploration of latent ontologies has yielded some surprising results: ELMo knows people, BERT knows dates, and both sense scalar and measurable values, while distinguishing between small and large numbers. Both models preferentially split core semantic roles into many fine-grained categories, and seem to encode broad notions of syntactic and semantic structure.
These findings contrast with those from fully-supervised probes, which produce strong agreement with existing annotations \citep{tenney2019what} but can also report false positives by fitting to weak patterns in large feature spaces \citep{zhang2018,voita2020informationtheoretic}.
Instead, agreement of latent categories with known concepts can be taken as strong evidence that these concepts (or similar ones) are present as important, salient features in an encoder's representation space.

This issue is particularly important when looking for \textit{deep, inherent understanding} of linguistic structure, which by nature must generalize.
For supervised systems, generalization
is often measured by out-of-distribution objectives like
out-of-domain performance \citep{ganin2016domain},
transferability \citep{wang2018glue},
targeted forms of compositionality \citep{geiger2020modular},
or robustness to adversarial inputs \citep{jia2017adversarial}.
Recent work also advocates for counterfactual learning and evaluation \citep{qin2019counterfactual,kaushik2020learning} to mitigate confounds,
or contrastive evaluation sets \cite{gardner2020evaluating} to rigorously test local decision boundaries.
Overall, these techniques target discrepancies between salient features in a model and causal relationships in a task.
In this work, we extract such features directly and investigate them by comparing induced and gold ontologies.
This identifies some very strong cases of transferability from the binary detection task to detection tasks over gold subcategories, such as ELMo's \textit{people} and BERT's \textit{dates} (\autoref{tab:ner-f1-results}).
Future work may investigate \textit{cross-task} ontology matching
to identify other transferable features,
the emergence of categories signifying pipelined reasoning \citep{tenney2019bert}, surface patterns, or new, perhaps unexpected distinctions which can appear when going beyond existing schemas \citep{michael2018crowdsourcing}.

Our results point to a paradigm of \textbf{probing with latent variables}, for which LSL is one potential technique.
We have only scratched the surface of what may emerge with such methods: while our probing test is high specificity, it is low power; extant latent structure may still be missed.
LSL probing may produce different ontologies due to many factors, such as
tokenization \citep{singh2019bert},
encoder architecture \citep{peters2018dissecting},
probe architecture \citep{hewitt-manning-2019-structural},
data distribution \citep{gururangan2018annotation},
pretraining task \citep{liu2019linguistic,wang2019tell},
or pretraining checkpoint.
Any such factors may be at work in the differences we observe between ELMo and BERT: for example, BERT's tokenization method may not as readily induce \textit{personhood} features due to splitting of rare words (like names) in byte-pair encoding.
Furthermore, concurrent work \citep{chi2020finding} has already found qualitative evidence of syntactic dependency types emergent in the special case of multilingual structural probes \citep{hewitt-manning-2019-structural}.
With LSL, we provide a method that can be adapted to a variety of probing settings to both quantify and qualify this kind of structure.



\section{Conclusion}

We introduced a new model analysis method based on
\textit{latent subclass learning}:
by factoring a binary classifier through a forced choice of latent subclasses,
latent ontologies can be coaxed out of input features.
Using this approach, we showed that encoders such as BERT and ELMo can be found to hold stable, consistent latent ontologies on a variety of linguistic tasks.
In these ontologies, we found clear connections to existing categories, such as \textit{personhood} of named entities.
We also found evidence of ontological distinctions beyond traditional gold categories,
such as distinguishing large and small numbers, or preferring fine-grained semantic roles for core arguments.
In latent subclass learning, we have shown a general technique to uncover some of these features discretely,
providing a starting point for descriptive analysis of our models' latent ontologies.
The high specificity of our method opens doors to more insights from future work, 
which may include investigating how LSL results vary with probe architecture, developing intrinsic quality measures on latent ontologies, or applying the technique to discover new patterns in settings where gold annotations are not present.

\section*{Acknowledgments}
We would like to thank Tim Dozat, Kenton Lee, Emily Pitler, Kellie Webster, other members of Google Research, Sewon Min, and the anonymous reviewers, who all provided valuable feedback on this paper. We also thank Rafael M\"{u}ller, Simon Kornblith, and Geoffrey Hinton for discussion on the LSL classifier, and Alessandro Sordoni for pointing out the connection between the clustering regularizers and mutual information.

\bibliographystyle{acl_natbib}
\bibliography{main}

\begin{thebibliography}{79}
\expandafter\ifx\csname natexlab\endcsname\relax\def\natexlab#1{#1}\fi

\bibitem[{Abnar et~al.(2019)Abnar, Beinborn, Choenni, and
  Zuidema}]{abnar2019blackbox}
Samira Abnar, Lisa Beinborn, Rochelle Choenni, and Willem Zuidema. 2019.
\newblock \href {https://doi.org/10.18653/v1/W19-4820} {Blackbox meets
  blackbox: Representational similarity and stability analysis of neural
  language models and brains}.
\newblock In \emph{Proceedings of the 2019 ACL Workshop BlackboxNLP: Analyzing
  and Interpreting Neural Networks for NLP}, pages 191--203.

\bibitem[{Adi et~al.(2017)Adi, Kermany, Belinkov, Lavi, and
  Goldberg}]{adi2016fine}
Yossi Adi, Einat Kermany, Yonatan Belinkov, Ofer Lavi, and Yoav Goldberg. 2017.
\newblock \href {https://openreview.net/forum?id=BJh6Ztuxl} {Fine-grained
  analysis of sentence embeddings using auxiliary prediction tasks}.
\newblock In \emph{International Conference on Learning Representations}.

\bibitem[{Amig{\'o} et~al.(2009)Amig{\'o}, Gonzalo, Artiles, and
  Verdejo}]{amigo2009comparison}
Enrique Amig{\'o}, Julio Gonzalo, Javier Artiles, and Felisa Verdejo. 2009.
\newblock A comparison of extrinsic clustering evaluation metrics based on
  formal constraints.
\newblock \emph{Information retrieval}, 12(4):461--486.

\bibitem[{Bagga and Baldwin(1998)}]{bagga-baldwin-1998-bcubed}
Amit Bagga and Breck Baldwin. 1998.
\newblock \href {https://doi.org/10.3115/980845.980859} {Entity-based
  cross-document coreferencing using the vector space model}.
\newblock In \emph{36th Annual Meeting of the Association for Computational
  Linguistics and 17th International Conference on Computational Linguistics,
  Volume 1}, pages 79--85, Montreal, Quebec, Canada. Association for
  Computational Linguistics.

\bibitem[{Baker et~al.(1998)Baker, Fillmore, and Lowe}]{baker1998berkeley}
Collin~F. Baker, Charles~J. Fillmore, and John~B. Lowe. 1998.
\newblock \href {https://www.aclweb.org/anthology/C98-1013} {The {B}erkeley
  {F}rame{N}et project}.
\newblock In \emph{{COLING} 1998 Volume 1: The 17th International Conference on
  Computational Linguistics}, pages 86--90. Association for Computational
  Linguistics.

\bibitem[{Banko et~al.(2007)Banko, Cafarella, Soderland, Broadhead, and
  Etzioni}]{banko2007open}
Michele Banko, Michael~J. Cafarella, Stephen Soderland, Matt Broadhead, and
  Oren Etzioni. 2007.
\newblock \href {http://dl.acm.org/citation.cfm?id=1625275.1625705} {Open
  information extraction from the web}.
\newblock In \emph{Proceedings of the 20th International Joint Conference on
  Artifical Intelligence}, IJCAI'07, pages 2670--2676, San Francisco, CA, USA.
  Morgan Kaufmann Publishers Inc.

\bibitem[{Belinkov et~al.(2017{\natexlab{a}})Belinkov, Durrani, Dalvi, Sajjad,
  and Glass}]{belinkov2017neural}
Yonatan Belinkov, Nadir Durrani, Fahim Dalvi, Hassan Sajjad, and James Glass.
  2017{\natexlab{a}}.
\newblock \href {https://doi.org/10.18653/v1/P17-1080} {What do neural machine
  translation models learn about morphology?}
\newblock In \emph{Proceedings of the 55th Annual Meeting of the Association
  for Computational Linguistics (Volume 1: Long Papers)}, pages 861--872.
  Association for Computational Linguistics.

\bibitem[{Belinkov and Glass(2019)}]{belinkov2019analysis}
Yonatan Belinkov and James Glass. 2019.
\newblock \href {https://doi.org/10.1162/tacl_a_00254} {Analysis methods in
  neural language processing: A survey}.
\newblock \emph{Transactions of the Association for Computational Linguistics},
  7:49--72.

\bibitem[{Belinkov et~al.(2017{\natexlab{b}})Belinkov, M{\`a}rquez, Sajjad,
  Durrani, Dalvi, and Glass}]{belinkov2017evaluating}
Yonatan Belinkov, Llu{\'\i}s M{\`a}rquez, Hassan Sajjad, Nadir Durrani, Fahim
  Dalvi, and James Glass. 2017{\natexlab{b}}.
\newblock \href {https://www.aclweb.org/anthology/I17-1001} {Evaluating layers
  of representation in neural machine translation on part-of-speech and
  semantic tagging tasks}.
\newblock In \emph{Proceedings of the Eighth International Joint Conference on
  Natural Language Processing (Volume 1: Long Papers)}, pages 1--10. Asian
  Federation of Natural Language Processing.

\bibitem[{Blevins et~al.(2018)Blevins, Levy, and
  Zettlemoyer}]{blevins2018hierarchical}
Terra Blevins, Omer Levy, and Luke Zettlemoyer. 2018.
\newblock \href {https://doi.org/10.18653/v1/P18-2003} {Deep {RNN}s encode soft
  hierarchical syntax}.
\newblock In \emph{Proceedings of the 56th Annual Meeting of the Association
  for Computational Linguistics (Volume 2: Short Papers)}, pages 14--19.
  Association for Computational Linguistics.

\bibitem[{Bouma(2009)}]{bouma2009normalized}
Gerlof Bouma. 2009.
\newblock Normalized (pointwise) mutual information in collocation extraction.
\newblock In \emph{GSCL}.

\bibitem[{Chelba et~al.(2014)Chelba, Mikolov, Schuster, Ge, Brants, Koehn, and
  Robinson}]{chelba2014one}
Ciprian Chelba, Tomas Mikolov, Mike Schuster, Qi~Ge, Thorsten Brants, Phillipp
  Koehn, and Tony Robinson. 2014.
\newblock One billion word benchmark for measuring progress in statistical
  language modeling.
\newblock In \emph{Proceedings of Interspeech}.

\bibitem[{Chen et~al.(2019)Chen, Chu, and Gimpel}]{chen-etal-2019-evaluation}
Mingda Chen, Zewei Chu, and Kevin Gimpel. 2019.
\newblock \href {https://doi.org/10.18653/v1/D19-1060} {Evaluation benchmarks
  and learning criteria for discourse-aware sentence representations}.
\newblock In \emph{Proceedings of the 2019 Conference on Empirical Methods in
  Natural Language Processing and the 9th International Joint Conference on
  Natural Language Processing (EMNLP-IJCNLP)}, pages 649--662, Hong Kong,
  China. Association for Computational Linguistics.

\bibitem[{Chi et~al.(2020)Chi, Hewitt, and Manning}]{chi2020finding}
Ethan~A. Chi, John Hewitt, and Christopher~D. Manning. 2020.
\newblock \href {https://doi.org/10.18653/v1/2020.acl-main.493} {Finding
  universal grammatical relations in multilingual {BERT}}.
\newblock In \emph{Proceedings of the 58th Annual Meeting of the Association
  for Computational Linguistics}, pages 5564--5577. Association for
  Computational Linguistics.

\bibitem[{Choi et~al.(2018)Choi, Levy, Choi, and Zettlemoyer}]{choi2018ultra}
Eunsol Choi, Omer Levy, Yejin Choi, and Luke Zettlemoyer. 2018.
\newblock \href {https://doi.org/10.18653/v1/P18-1009} {Ultra-fine entity
  typing}.
\newblock In \emph{Proceedings of the 56th Annual Meeting of the Association
  for Computational Linguistics (Volume 1: Long Papers)}, pages 87--96,
  Melbourne, Australia. Association for Computational Linguistics.

\bibitem[{Chrupa{\l}a and Alishahi(2019)}]{chrupala2019correlating}
Grzegorz Chrupa{\l}a and Afra Alishahi. 2019.
\newblock \href {https://doi.org/10.18653/v1/P19-1283} {Correlating neural and
  symbolic representations of language}.
\newblock In \emph{Proceedings of the 57th Annual Meeting of the Association
  for Computational Linguistics}, pages 2952--2962, Florence, Italy.
  Association for Computational Linguistics.

\bibitem[{Church and Hanks(1989)}]{church1990word}
Kenneth~Ward Church and Patrick Hanks. 1989.
\newblock \href {https://doi.org/10.3115/981623.981633} {Word association
  norms, mutual information, and lexicography}.
\newblock In \emph{27th Annual Meeting of the Association for Computational
  Linguistics}, volume~16, pages 76--83. Association for Computational
  Linguistics.

\bibitem[{Clark et~al.(2019)Clark, Khandelwal, Levy, and
  Manning}]{clark2019what}
Kevin Clark, Urvashi Khandelwal, Omer Levy, and Christopher~D. Manning. 2019.
\newblock \href {https://doi.org/10.18653/v1/W19-4828} {What does {BERT} look
  at? an analysis of {BERT}{'}s attention}.
\newblock In \emph{Proceedings of the 2019 ACL Workshop BlackboxNLP: Analyzing
  and Interpreting Neural Networks for NLP}, pages 276--286. Association for
  Computational Linguistics.

\bibitem[{Conneau et~al.(2018)Conneau, Kruszewski, Lample, Barrault, and
  Baroni}]{conneau2018cram}
Alexis Conneau, German Kruszewski, Guillaume Lample, Lo{\"\i}c Barrault, and
  Marco Baroni. 2018.
\newblock \href {https://doi.org/10.18653/v1/P18-1198} {What you can cram into
  a single {\$}{\&}!{\#}* vector: Probing sentence embeddings for linguistic
  properties}.
\newblock In \emph{Proceedings of the 56th Annual Meeting of the Association
  for Computational Linguistics (Volume 1: Long Papers)}, pages 2126--2136.
  Association for Computational Linguistics.

\bibitem[{Devlin et~al.(2019)Devlin, Chang, Lee, and
  Toutanova}]{devlin2019bert}
Jacob Devlin, Ming-Wei Chang, Kenton Lee, and Kristina Toutanova. 2019.
\newblock \href {https://doi.org/10.18653/v1/N19-1423} {{BERT}: Pre-training of
  deep bidirectional transformers for language understanding}.
\newblock In \emph{Proceedings of the 2019 Conference of the North {A}merican
  Chapter of the Association for Computational Linguistics: Human Language
  Technologies, Volume 1 (Long and Short Papers)}, pages 4171--4186,
  Minneapolis, Minnesota. Association for Computational Linguistics.

\bibitem[{Dowty(1991)}]{dowty1991thematic}
David Dowty. 1991.
\newblock Thematic proto-roles and argument selection.
\newblock \emph{Language}, 67:547--619.

\bibitem[{Fillmore et~al.(2006)}]{fillmore2006frame}
Charles~J Fillmore et~al. 2006.
\newblock Frame semantics.
\newblock \emph{Cognitive linguistics: Basic readings}, 34:373--400.

\bibitem[{FitzGerald et~al.(2018)FitzGerald, Michael, He, and
  Zettlemoyer}]{fitzgerald2018large}
Nicholas FitzGerald, Julian Michael, Luheng He, and Luke Zettlemoyer. 2018.
\newblock \href {https://doi.org/10.18653/v1/P18-1191} {Large-scale {QA}-{SRL}
  parsing}.
\newblock In \emph{Proceedings of the 56th Annual Meeting of the Association
  for Computational Linguistics (Volume 1: Long Papers)}, pages 2051--2060.
  Association for Computational Linguistics.

\bibitem[{Ganin et~al.(2016)Ganin, Ustinova, Ajakan, Germain, Larochelle,
  Laviolette, Marchand, and Lempitsky}]{ganin2016domain}
Yaroslav Ganin, Evgeniya Ustinova, Hana Ajakan, Pascal Germain, Hugo
  Larochelle, Fran{\c{c}}ois Laviolette, Mario Marchand, and Victor Lempitsky.
  2016.
\newblock Domain-adversarial training of neural networks.
\newblock \emph{The Journal of Machine Learning Research}, 17(1):2096--2030.

\bibitem[{Gardner et~al.(2020)Gardner, Artzi, Basmova, Berant, Bogin, Chen,
  Dasigi, Dua, Elazar, Gottumukkala, Gupta, Hajishirzi, Ilharco, Khashabi, Lin,
  Liu, Liu, Mulcaire, Ning, Singh, Smith, Subramanian, Tsarfaty, Wallace,
  Zhang, and Zhou}]{gardner2020evaluating}
Matt Gardner, Yoav Artzi, Victoria Basmova, Jonathan Berant, Ben Bogin, Sihao
  Chen, Pradeep Dasigi, Dheeru Dua, Yanai Elazar, Ananth Gottumukkala, Nitish
  Gupta, Hanna Hajishirzi, Gabriel Ilharco, Daniel Khashabi, Kevin Lin,
  Jiangming Liu, Nelson~F. Liu, Phoebe Mulcaire, Qiang Ning, Sameer Singh,
  Noah~A. Smith, Sanjay Subramanian, Reut Tsarfaty, Eric Wallace, Ally Zhang,
  and Ben Zhou. 2020.
\newblock \href {http://arxiv.org/abs/2004.02709} {Evaluating {NLP} models via
  contrast sets}.

\bibitem[{Geiger et~al.(2020)Geiger, Richardson, and Potts}]{geiger2020modular}
Atticus Geiger, Kyle Richardson, and Christopher Potts. 2020.
\newblock Modular representation underlies systematic generalization in neural
  natural language inference models.
\newblock \emph{arXiv preprint arXiv:2004.14623}.

\bibitem[{Goldberg(2019)}]{goldberg2019assessing}
Yoav Goldberg. 2019.
\newblock Assessing bert's syntactic abilities.
\newblock \emph{arXiv preprint arXiv:1901.05287}.

\bibitem[{Gururangan et~al.(2018)Gururangan, Swayamdipta, Levy, Schwartz,
  Bowman, and Smith}]{gururangan2018annotation}
Suchin Gururangan, Swabha Swayamdipta, Omer Levy, Roy Schwartz, Samuel Bowman,
  and Noah~A. Smith. 2018.
\newblock \href {https://doi.org/10.18653/v1/N18-2017} {Annotation artifacts in
  natural language inference data}.
\newblock In \emph{Proceedings of the 2018 Conference of the North {A}merican
  Chapter of the Association for Computational Linguistics: Human Language
  Technologies, Volume 2 (Short Papers)}, volume~2, pages 107--112. Association
  for Computational Linguistics.

\bibitem[{He et~al.(2017)He, Lee, Lewis, and Zettlemoyer}]{he2017deep}
Luheng He, Kenton Lee, Mike Lewis, and Luke Zettlemoyer. 2017.
\newblock \href {https://doi.org/10.18653/v1/P17-1044} {Deep semantic role
  labeling: What works and what{'}s next}.
\newblock In \emph{Proceedings of the 55th Annual Meeting of the Association
  for Computational Linguistics (Volume 1: Long Papers)}, pages 473--483,
  Vancouver, Canada. Association for Computational Linguistics.

\bibitem[{He et~al.(2015)He, Lewis, and Zettlemoyer}]{he2015question}
Luheng He, Mike Lewis, and Luke Zettlemoyer. 2015.
\newblock \href {https://doi.org/10.18653/v1/D15-1076} {Question-answer driven
  semantic role labeling: Using natural language to annotate natural language}.
\newblock In \emph{Proceedings of the 2015 Conference on Empirical Methods in
  Natural Language Processing}, pages 643--653. Association for Computational
  Linguistics.

\bibitem[{Hewitt and Liang(2019)}]{hewitt-liang-2019-designing}
John Hewitt and Percy Liang. 2019.
\newblock \href {https://doi.org/10.18653/v1/D19-1275} {Designing and
  interpreting probes with control tasks}.
\newblock In \emph{Proceedings of the 2019 Conference on Empirical Methods in
  Natural Language Processing and the 9th International Joint Conference on
  Natural Language Processing (EMNLP-IJCNLP)}, pages 2733--2743, Hong Kong,
  China. Association for Computational Linguistics.

\bibitem[{Hewitt and Manning(2019)}]{hewitt-manning-2019-structural}
John Hewitt and Christopher~D. Manning. 2019.
\newblock \href {https://doi.org/10.18653/v1/N19-1419} {{A} structural probe
  for finding syntax in word representations}.
\newblock In \emph{Proceedings of the 2019 Conference of the North {A}merican
  Chapter of the Association for Computational Linguistics: Human Language
  Technologies, Volume 1 (Long and Short Papers)}, pages 4129--4138,
  Minneapolis, Minnesota. Association for Computational Linguistics.

\bibitem[{Hochreiter and Schmidhuber(1997)}]{hochreiter1997long}
Sepp Hochreiter and J{\"u}rgen Schmidhuber. 1997.
\newblock Long short-term memory.
\newblock \emph{Neural computation}, 9(8):1735--1780.

\bibitem[{Jia and Liang(2017)}]{jia2017adversarial}
Robin Jia and Percy Liang. 2017.
\newblock \href {https://doi.org/10.18653/v1/D17-1215} {Adversarial examples
  for evaluating reading comprehension systems}.
\newblock In \emph{Proceedings of the 2017 Conference on Empirical Methods in
  Natural Language Processing}, pages 2021--2031, Copenhagen, Denmark.
  Association for Computational Linguistics.

\bibitem[{Kaushik et~al.(2020)Kaushik, Hovy, and Lipton}]{kaushik2020learning}
Divyansh Kaushik, Eduard Hovy, and Zachary Lipton. 2020.
\newblock \href {https://openreview.net/forum?id=Sklgs0NFvr} {Learning the
  difference that makes a difference with counterfactually-augmented data}.
\newblock In \emph{International Conference on Learning Representations}.

\bibitem[{Kingsbury et~al.(2002)Kingsbury, Palmer, and
  Marcus}]{kingsbury2002adding}
Paul Kingsbury, Martha Palmer, and Mitch Marcus. 2002.
\newblock Adding semantic annotation to the penn treebank.
\newblock In \emph{In Proceedings of the Human Language Technology Conference}.

\bibitem[{Kuznetsov and Gurevych(2020)}]{kuznetsov2020matter}
Ilia Kuznetsov and Iryna Gurevych. 2020.
\newblock A matter of framing: The impact of linguistic formalism on probing
  results.
\newblock \emph{arXiv preprint arXiv:2004.14999}.

\bibitem[{Lan et~al.(2020)Lan, Chen, Goodman, Gimpel, Sharma, and
  Soricut}]{lan2020albert}
Zhenzhong Lan, Mingda Chen, Sebastian Goodman, Kevin Gimpel, Piyush Sharma, and
  Radu Soricut. 2020.
\newblock \href {https://openreview.net/forum?id=H1eA7AEtvS} {Albert: A lite
  bert for self-supervised learning of language representations}.
\newblock In \emph{International Conference on Learning Representations}.

\bibitem[{Lee et~al.(2017)Lee, He, Lewis, and Zettlemoyer}]{lee2017end}
Kenton Lee, Luheng He, Mike Lewis, and Luke Zettlemoyer. 2017.
\newblock \href {https://doi.org/10.18653/v1/D17-1018} {End-to-end neural
  coreference resolution}.
\newblock In \emph{Proceedings of the 2017 Conference on Empirical Methods in
  Natural Language Processing}, pages 188--197. Association for Computational
  Linguistics.

\bibitem[{Levin(1993)}]{levin1993english}
B.~Levin. 1993.
\newblock \href {https://books.google.com/books?id=6wIZWOrcBf8C} {\emph{English
  Verb Classes and Alternations: A Preliminary Investigation}}.
\newblock University of Chicago Press.

\bibitem[{Ling and Weld(2012)}]{ling2012finegrained}
Xiao Ling and Daniel~S. Weld. 2012.
\newblock \href {http://dl.acm.org/citation.cfm?id=2900728.2900742}
  {Fine-grained entity recognition}.
\newblock In \emph{Proceedings of the Twenty-Sixth AAAI Conference on
  Artificial Intelligence}, AAAI'12, pages 94--100. AAAI Press.

\bibitem[{Liu et~al.(2019{\natexlab{a}})Liu, Gardner, Belinkov, Peters, and
  Smith}]{liu2019linguistic}
Nelson~F. Liu, Matt Gardner, Yonatan Belinkov, Matthew~E. Peters, and Noah~A.
  Smith. 2019{\natexlab{a}}.
\newblock \href {https://doi.org/10.18653/v1/N19-1112} {Linguistic knowledge
  and transferability of contextual representations}.
\newblock In \emph{Proceedings of the 2019 Conference of the North {A}merican
  Chapter of the Association for Computational Linguistics: Human Language
  Technologies, Volume 1 (Long and Short Papers)}, pages 1073--1094,
  Minneapolis, Minnesota. Association for Computational Linguistics.

\bibitem[{Liu et~al.(2019{\natexlab{b}})Liu, He, Chen, and Gao}]{liu2019multi}
Xiaodong Liu, Pengcheng He, Weizhu Chen, and Jianfeng Gao. 2019{\natexlab{b}}.
\newblock \href {https://doi.org/10.18653/v1/P19-1441} {Multi-task deep neural
  networks for natural language understanding}.
\newblock In \emph{Proceedings of the 57th Annual Meeting of the Association
  for Computational Linguistics}, pages 4487--4496, Florence, Italy.
  Association for Computational Linguistics.

\bibitem[{Marcus et~al.(1993)Marcus, Santorini, and
  Marcinkiewicz}]{marcus1993building}
Mitchell~P. Marcus, Beatrice Santorini, and Mary~Ann Marcinkiewicz. 1993.
\newblock \href {https://www.aclweb.org/anthology/J93-2004} {Building a large
  annotated corpus of {E}nglish: The {P}enn {T}reebank}.
\newblock \emph{Computational Linguistics}, 19(2):313--330.

\bibitem[{Marvin and Linzen(2018)}]{marvin2018targeted}
Rebecca Marvin and Tal Linzen. 2018.
\newblock \href {https://doi.org/10.18653/v1/D18-1151} {Targeted syntactic
  evaluation of language models}.
\newblock In \emph{Proceedings of the 2018 Conference on Empirical Methods in
  Natural Language Processing}, pages 1192--1202. Association for Computational
  Linguistics.

\bibitem[{McInnes et~al.(2018)McInnes, Healy, and Melville}]{mcinnes2018umap}
Leland McInnes, John Healy, and James Melville. 2018.
\newblock Umap: Uniform manifold approximation and projection for dimension
  reduction.
\newblock \emph{arXiv preprint arXiv:1802.03426}.

\bibitem[{Michael et~al.(2018)Michael, Stanovsky, He, Dagan, and
  Zettlemoyer}]{michael2018crowdsourcing}
Julian Michael, Gabriel Stanovsky, Luheng He, Ido Dagan, and Luke Zettlemoyer.
  2018.
\newblock \href {https://doi.org/10.18653/v1/N18-2089} {Crowdsourcing
  question-answer meaning representations}.
\newblock In \emph{Proceedings of the 2018 Conference of the North {A}merican
  Chapter of the Association for Computational Linguistics: Human Language
  Technologies, Volume 2 (Short Papers)}, pages 560--568. Association for
  Computational Linguistics.

\bibitem[{Müller et~al.(2020)Müller, Kornblith, and
  Hinton}]{muller2020subclass}
Rafael Müller, Simon Kornblith, and Geoffrey Hinton. 2020.
\newblock Subclass distillation.
\newblock \emph{arXiv preprint arXiv:2002.03936}.

\bibitem[{Nivre et~al.(2015)Nivre, Agi{\'c}, Aranzabe, Asahara, Atutxa,
  Ballesteros, Bauer, Bengoetxea, Bhat, Bosco, Bowman, Celano, Connor,
  de~Marneffe, Diaz~de Ilarraza, Dobrovoljc, Dozat, Erjavec, Farkas, Foster,
  Galbraith, Ginter, Goenaga, Gojenola, Goldberg, Gonzales, Guillaume, Haji{\v
  c}, Haug, Ion, Irimia, Johannsen, Kanayama, Kanerva, Krek, Laippala, Lenci,
  Ljube{\v s}i{\'c}, Lynn, Manning, M{\u a}r{\u a}nduc, Mare{\v c}ek,
  Mart{\'{\i}}nez~Alonso, Ma{\v s}ek, Matsumoto, {McDonald}, Missil{\"a},
  Mititelu, Miyao, Montemagni, Mori, Nurmi, Osenova, {\O}vrelid, Pascual,
  Passarotti, Perez, Petrov, Piitulainen, Plank, Popel, Prokopidis, Pyysalo,
  Ramasamy, Rosa, Saleh, Schuster, Seeker, Seraji, Silveira, Simi, Simionescu,
  Simk{\'o}, Simov, Smith, {\v S}t{\v e}p{\'a}nek, Suhr, Sz{\'a}nt{\'o},
  Tanaka, Tsarfaty, Uematsu, Uria, Varga, Vincze, {\v Z}abokrtsk{\'y}, Zeman,
  and Zhu}]{nivre2015universal}
Joakim Nivre, {\v Z}eljko Agi{\'c}, Maria~Jesus Aranzabe, Masayuki Asahara,
  Aitziber Atutxa, Miguel Ballesteros, John Bauer, Kepa Bengoetxea, Riyaz~Ahmad
  Bhat, Cristina Bosco, Sam Bowman, Giuseppe G.~A. Celano, Miriam Connor,
  Marie-Catherine de~Marneffe, Arantza Diaz~de Ilarraza, Kaja Dobrovoljc,
  Timothy Dozat, Toma{\v z} Erjavec, Rich{\'a}rd Farkas, Jennifer Foster,
  Daniel Galbraith, Filip Ginter, Iakes Goenaga, Koldo Gojenola, Yoav Goldberg,
  Berta Gonzales, Bruno Guillaume, Jan Haji{\v c}, Dag Haug, Radu Ion, Elena
  Irimia, Anders Johannsen, Hiroshi Kanayama, Jenna Kanerva, Simon Krek,
  Veronika Laippala, Alessandro Lenci, Nikola Ljube{\v s}i{\'c}, Teresa Lynn,
  Christopher Manning, C{\u a}t{\u a}lina M{\u a}r{\u a}nduc, David Mare{\v
  c}ek, H{\'e}ctor Mart{\'{\i}}nez~Alonso, Jan Ma{\v s}ek, Yuji Matsumoto, Ryan
  {McDonald}, Anna Missil{\"a}, Verginica Mititelu, Yusuke Miyao, Simonetta
  Montemagni, Shunsuke Mori, Hanna Nurmi, Petya Osenova, Lilja {\O}vrelid,
  Elena Pascual, Marco Passarotti, Cenel-Augusto Perez, Slav Petrov, Jussi
  Piitulainen, Barbara Plank, Martin Popel, Prokopis Prokopidis, Sampo Pyysalo,
  Loganathan Ramasamy, Rudolf Rosa, Shadi Saleh, Sebastian Schuster, Wolfgang
  Seeker, Mojgan Seraji, Natalia Silveira, Maria Simi, Radu Simionescu, Katalin
  Simk{\'o}, Kiril Simov, Aaron Smith, Jan {\v S}t{\v e}p{\'a}nek, Alane Suhr,
  Zsolt Sz{\'a}nt{\'o}, Takaaki Tanaka, Reut Tsarfaty, Sumire Uematsu, Larraitz
  Uria, Viktor Varga, Veronika Vincze, Zden{\v e}k {\v Z}abokrtsk{\'y}, Daniel
  Zeman, and Hanzhi Zhu. 2015.
\newblock \href {http://hdl.handle.net/11234/1-1548} {Universal dependencies
  1.2}.
\newblock {LINDAT}/{CLARIAH}-{CZ} digital library at the Institute of Formal
  and Applied Linguistics ({{\'U}FAL}), Faculty of Mathematics and Physics,
  Charles University.

\bibitem[{Palmer et~al.(2005)Palmer, Gildea, and
  Kingsbury}]{palmer2005proposition}
Martha Palmer, Daniel Gildea, and Paul Kingsbury. 2005.
\newblock \href {https://doi.org/10.1162/0891201053630264} {The {P}roposition
  {B}ank: An annotated corpus of semantic roles}.
\newblock \emph{Computational Linguistics}, 31(1):71--106.

\bibitem[{Peters et~al.(2018{\natexlab{a}})Peters, Neumann, Iyyer, Gardner,
  Clark, Lee, and Zettlemoyer}]{peters2018deep}
Matthew Peters, Mark Neumann, Mohit Iyyer, Matt Gardner, Christopher Clark,
  Kenton Lee, and Luke Zettlemoyer. 2018{\natexlab{a}}.
\newblock \href {https://doi.org/10.18653/v1/N18-1202} {Deep contextualized
  word representations}.
\newblock In \emph{Proceedings of the 2018 Conference of the North {A}merican
  Chapter of the Association for Computational Linguistics: Human Language
  Technologies, Volume 1 (Long Papers)}, pages 2227--2237. Association for
  Computational Linguistics.

\bibitem[{Peters et~al.(2018{\natexlab{b}})Peters, Neumann, Zettlemoyer, and
  Yih}]{peters2018dissecting}
Matthew Peters, Mark Neumann, Luke Zettlemoyer, and Wen-tau Yih.
  2018{\natexlab{b}}.
\newblock \href {https://doi.org/10.18653/v1/D18-1179} {Dissecting contextual
  word embeddings: Architecture and representation}.
\newblock In \emph{Proceedings of the 2018 Conference on Empirical Methods in
  Natural Language Processing}, pages 1499--1509. Association for Computational
  Linguistics.

\bibitem[{Petroni et~al.(2019)Petroni, Rockt{\"a}schel, Riedel, Lewis, Bakhtin,
  Wu, and Miller}]{petroni2019language}
Fabio Petroni, Tim Rockt{\"a}schel, Sebastian Riedel, Patrick Lewis, Anton
  Bakhtin, Yuxiang Wu, and Alexander Miller. 2019.
\newblock \href {https://doi.org/10.18653/v1/D19-1250} {Language models as
  knowledge bases?}
\newblock In \emph{Proceedings of the 2019 Conference on Empirical Methods in
  Natural Language Processing and the 9th International Joint Conference on
  Natural Language Processing (EMNLP-IJCNLP)}, pages 2463--2473, Hong Kong,
  China. Association for Computational Linguistics.

\bibitem[{Pimentel et~al.(2020)Pimentel, Valvoda, Hall~Maudslay, Zmigrod,
  Williams, and Cotterell}]{pimentel2020informationtheoretic}
Tiago Pimentel, Josef Valvoda, Rowan Hall~Maudslay, Ran Zmigrod, Adina
  Williams, and Ryan Cotterell. 2020.
\newblock \href {https://doi.org/10.18653/v1/2020.acl-main.420}
  {Information-theoretic probing for linguistic structure}.
\newblock In \emph{Proceedings of the 58th Annual Meeting of the Association
  for Computational Linguistics}, pages 4609--4622. Association for
  Computational Linguistics, Association for Computational Linguistics.

\bibitem[{Pradhan et~al.(2007)Pradhan, Hovy, Marcus, Palmer, Ramshaw, and
  Weischedel}]{pradhan2007ontonotes}
Sameer~S Pradhan, Eduard Hovy, Mitch Marcus, Martha Palmer, Lance Ramshaw, and
  Ralph Weischedel. 2007.
\newblock Ontonotes: A unified relational semantic representation.
\newblock \emph{International Journal of Semantic Computing}, 1(04):405--419.

\bibitem[{Qin et~al.(2019)Qin, Bosselut, Holtzman, Bhagavatula, Clark, and
  Choi}]{qin2019counterfactual}
Lianhui Qin, Antoine Bosselut, Ari Holtzman, Chandra Bhagavatula, Elizabeth
  Clark, and Yejin Choi. 2019.
\newblock \href {https://doi.org/10.18653/v1/D19-1509} {Counterfactual story
  reasoning and generation}.
\newblock In \emph{Proceedings of the 2019 Conference on Empirical Methods in
  Natural Language Processing and the 9th International Joint Conference on
  Natural Language Processing (EMNLP-IJCNLP)}, pages 5043--5053, Hong Kong,
  China. Association for Computational Linguistics.

\bibitem[{Radford et~al.(2019)Radford, Wu, Child, Luan, Amodei, and
  Sutskever}]{radford2019language}
Alec Radford, Jeffrey Wu, Rewon Child, David Luan, Dario Amodei, and Ilya
  Sutskever. 2019.
\newblock Language models are unsupervised multitask learners.
\newblock \emph{https://blog.openai.com/better-language-models}.

\bibitem[{Reif et~al.(2019)Reif, Yuan, Wattenberg, Viegas, Coenen, Pearce, and
  Kim}]{reif2019bertviz}
Emily Reif, Ann Yuan, Martin Wattenberg, Fernanda~B Viegas, Andy Coenen, Adam
  Pearce, and Been Kim. 2019.
\newblock \href
  {http://papers.nips.cc/paper/9065-visualizing-and-measuring-the-geometry-of-bert.pdf}
  {Visualizing and measuring the geometry of {BERT}}.
\newblock In H.~Wallach, H.~Larochelle, A.~Beygelzimer, F.~d\textquotesingle
  Alch\'{e}-Buc, E.~Fox, and R.~Garnett, editors, \emph{Advances in Neural
  Information Processing Systems 32}, pages 8592--8600. Curran Associates, Inc.

\bibitem[{Reisinger et~al.(2015)Reisinger, Rudinger, Ferraro, Harman, Rawlins,
  and Van~Durme}]{reisinger2015semantic}
Drew Reisinger, Rachel Rudinger, Francis Ferraro, Craig Harman, Kyle Rawlins,
  and Benjamin Van~Durme. 2015.
\newblock \href {https://doi.org/10.1162/tacl_a_00152} {Semantic proto-roles}.
\newblock \emph{Transactions of the Association for Computational Linguistics},
  pages 475--488.

\bibitem[{Rogers et~al.(2020)Rogers, Kovaleva, and
  Rumshisky}]{rogers2020primer}
Anna Rogers, Olga Kovaleva, and Anna Rumshisky. 2020.
\newblock A primer in {BERTology}: What we know about how {BERT} works.
\newblock \emph{arXiv preprint arXiv:2002.12327}.

\bibitem[{Saphra and Lopez(2019)}]{saphra-lopez-2019-understanding}
Naomi Saphra and Adam Lopez. 2019.
\newblock \href {https://doi.org/10.18653/v1/N19-1329} {Understanding learning
  dynamics of language models with {SVCCA}}.
\newblock In \emph{Proceedings of the 2019 Conference of the North {A}merican
  Chapter of the Association for Computational Linguistics: Human Language
  Technologies, Volume 1 (Long and Short Papers)}, pages 3257--3267,
  Minneapolis, Minnesota. Association for Computational Linguistics.

\bibitem[{Schuler(2005)}]{schuler2005verbnet}
Karin~Kipper Schuler. 2005.
\newblock \emph{VerbNet: A broad-coverage, comprehensive verb lexicon}.
\newblock Ph.D. thesis, University of Pennsylvania.

\bibitem[{Silveira et~al.(2014)Silveira, Dozat, de~Marneffe, Bowman, Connor,
  Bauer, and Manning}]{silveira14gold}
Natalia Silveira, Timothy Dozat, Marie-Catherine de~Marneffe, Samuel Bowman,
  Miriam Connor, John Bauer, and Chris Manning. 2014.
\newblock \href
  {http://www.lrec-conf.org/proceedings/lrec2014/pdf/1089_Paper.pdf} {A gold
  standard dependency corpus for {E}nglish}.
\newblock In \emph{Proceedings of the Ninth International Conference on
  Language Resources and Evaluation ({LREC}'14)}, pages 2897--2904. European
  Language Resources Association (ELRA).

\bibitem[{Singh et~al.(2019)Singh, McCann, Socher, and Xiong}]{singh2019bert}
Jasdeep Singh, Bryan McCann, Richard Socher, and Caiming Xiong. 2019.
\newblock \href {https://doi.org/10.18653/v1/D19-6106} {{BERT} is not an
  interlingua and the bias of tokenization}.
\newblock In \emph{Proceedings of the 2nd Workshop on Deep Learning Approaches
  for Low-Resource NLP (DeepLo 2019)}, pages 47--55, Hong Kong, China.
  Association for Computational Linguistics.

\bibitem[{Smilkov et~al.(2016)Smilkov, Thorat, Nicholson, Reif, Vi{\'e}gas, and
  Wattenberg}]{smilkov2016embedding}
Daniel Smilkov, Nikhil Thorat, Charles Nicholson, Emily Reif, Fernanda~B
  Vi{\'e}gas, and Martin Wattenberg. 2016.
\newblock Embedding projector: Interactive visualization and interpretation of
  embeddings.
\newblock In \emph{NIPS 2016 Workshop on Interpretable Machine Learning in
  Complex Systems}.

\bibitem[{Talmor et~al.(2019)Talmor, Elazar, Goldberg, and
  Berant}]{talmor2019olmpics}
Alon Talmor, Yanai Elazar, Yoav Goldberg, and Jonathan Berant. 2019.
\newblock {oLMpics} -- on what language model pre-training captures.
\newblock \emph{arXiv preprint arXiv:1912.13283}.

\bibitem[{Tenney et~al.(2019{\natexlab{a}})Tenney, Das, and
  Pavlick}]{tenney2019bert}
Ian Tenney, Dipanjan Das, and Ellie Pavlick. 2019{\natexlab{a}}.
\newblock \href {https://doi.org/10.18653/v1/P19-1452} {{BERT} rediscovers the
  classical {NLP} pipeline}.
\newblock In \emph{Proceedings of the 57th Annual Meeting of the Association
  for Computational Linguistics}, pages 4593--4601, Florence, Italy.
  Association for Computational Linguistics.

\bibitem[{Tenney et~al.(2019{\natexlab{b}})Tenney, Xia, Chen, Wang, Poliak,
  McCoy, Kim, Durme, Bowman, Das, and Pavlick}]{tenney2019what}
Ian Tenney, Patrick Xia, Berlin Chen, Alex Wang, Adam Poliak, R~Thomas McCoy,
  Najoung Kim, Benjamin~Van Durme, Sam Bowman, Dipanjan Das, and Ellie Pavlick.
  2019{\natexlab{b}}.
\newblock What do you learn from context? probing for sentence structure in
  contextualized word representations.
\newblock In \emph{International Conference on Learning Representations}.

\bibitem[{Vaswani et~al.(2017)Vaswani, Shazeer, Parmar, Uszkoreit, Jones,
  Gomez, Kaiser, and Polosukhin}]{vaswani2017attention}
Ashish Vaswani, Noam Shazeer, Niki Parmar, Jakob Uszkoreit, Llion Jones,
  Aidan~N Gomez, {\L}ukasz Kaiser, and Illia Polosukhin. 2017.
\newblock Attention is all you need.
\newblock In \emph{Proceedings of NIPS}.

\bibitem[{Voita et~al.(2019)Voita, Sennrich, and Titov}]{voita2019bottom}
Elena Voita, Rico Sennrich, and Ivan Titov. 2019.
\newblock \href {https://doi.org/10.18653/v1/D19-1448} {The bottom-up evolution
  of representations in the transformer: A study with machine translation and
  language modeling objectives}.
\newblock In \emph{Proceedings of the 2019 Conference on Empirical Methods in
  Natural Language Processing and the 9th International Joint Conference on
  Natural Language Processing (EMNLP-IJCNLP)}, pages 4396--4406. Association
  for Computational Linguistics.

\bibitem[{Voita and Titov(2020)}]{voita2020informationtheoretic}
Elena Voita and Ivan Titov. 2020.
\newblock Information-theoretic probing with minimum description length.
\newblock \emph{arXiv preprint arXiv:2003.12298}.

\bibitem[{Wang et~al.(2019{\natexlab{a}})Wang, Hula, Xia, Pappagari, McCoy,
  Patel, Kim, Tenney, Huang, Yu, Jin, Chen, Van~Durme, Grave, Pavlick, and
  Bowman}]{wang2019tell}
Alex Wang, Jan Hula, Patrick Xia, Raghavendra Pappagari, R.~Thomas McCoy, Roma
  Patel, Najoung Kim, Ian Tenney, Yinghui Huang, Katherin Yu, Shuning Jin,
  Berlin Chen, Benjamin Van~Durme, Edouard Grave, Ellie Pavlick, and Samuel~R.
  Bowman. 2019{\natexlab{a}}.
\newblock \href {https://doi.org/10.18653/v1/P19-1439} {Can you tell me how to
  get past sesame street? sentence-level pretraining beyond language modeling}.
\newblock In \emph{Proceedings of the 57th Annual Meeting of the Association
  for Computational Linguistics}, pages 4465--4476, Florence, Italy.
  Association for Computational Linguistics.

\bibitem[{Wang et~al.(2019{\natexlab{b}})Wang, Pruksachatkun, Nangia, Singh,
  Michael, Hill, Levy, and Bowman}]{wang2019superglue}
Alex Wang, Yada Pruksachatkun, Nikita Nangia, Amanpreet Singh, Julian Michael,
  Felix Hill, Omer Levy, and Samuel~R. Bowman. 2019{\natexlab{b}}.
\newblock \href
  {http://papers.nips.cc/paper/8589-superglue-a-stickier-benchmark-for-general-purpose-language-understanding-systems.pdf}
  {{S}uper{GLUE}: A multi-task benchmark and analysis platform for natural
  language understanding}.
\newblock In H.~Wallach, H.~Larochelle, A.~Beygelzimer, F.~d\textquotesingle
  Alch\'{e}-Buc, E.~Fox, and R.~Garnett, editors, \emph{Advances in Neural
  Information Processing Systems 32}, pages 3261--3275. Curran Associates, Inc.

\bibitem[{Wang et~al.(2018)Wang, Singh, Michael, Hill, Levy, and
  Bowman}]{wang2018glue}
Alex Wang, Amanpreet Singh, Julian Michael, Felix Hill, Omer Levy, and Samuel
  Bowman. 2018.
\newblock \href {https://doi.org/10.18653/v1/W18-5446} {{GLUE}: A multi-task
  benchmark and analysis platform for natural language understanding}.
\newblock In \emph{Proceedings of the 2018 {EMNLP} Workshop {B}lackbox{NLP}:
  Analyzing and Interpreting Neural Networks for {NLP}}, pages 353--355.
  Association for Computational Linguistics.

\bibitem[{Weischedel et~al.(2013)Weischedel, Palmer, Marcus, Hovy, Pradhan,
  Ramshaw, Xue, Taylor, Kaufman, Franchini et~al.}]{weischedel2013ontonotes}
Ralph Weischedel, Martha Palmer, Mitchell Marcus, Eduard Hovy, Sameer Pradhan,
  Lance Ramshaw, Nianwen Xue, Ann Taylor, Jeff Kaufman, Michelle Franchini,
  et~al. 2013.
\newblock {OntoNotes} release 5.0 {LDC2013T19}.
\newblock \emph{Linguistic Data Consortium, Philadelphia, PA}.

\bibitem[{Wu et~al.(2020)Wu, Chen, Kao, and Liu}]{wu2020perturbed}
Zhiyong Wu, Yun Chen, Ben Kao, and Qun Liu. 2020.
\newblock \href {https://doi.org/10.18653/v1/2020.acl-main.383} {Perturbed
  masking: Parameter-free probing for analyzing and interpreting {BERT}}.
\newblock In \emph{Proceedings of the 58th Annual Meeting of the Association
  for Computational Linguistics}, pages 4166--4176. Association for
  Computational Linguistics.

\bibitem[{Zhang and Bowman(2018)}]{zhang2018}
Kelly Zhang and Samuel Bowman. 2018.
\newblock \href {https://doi.org/10.18653/v1/W18-5448} {Language modeling
  teaches you more than translation does: Lessons learned through auxiliary
  syntactic task analysis}.
\newblock In \emph{Proceedings of the 2018 {EMNLP} Workshop {B}lackbox{NLP}:
  Analyzing and Interpreting Neural Networks for {NLP}}, pages 359--361.
  Association for Computational Linguistics.

\bibitem[{Zhang et~al.(2017)Zhang, Zhong, Chen, Angeli, and
  Manning}]{zhang2017tacred}
Yuhao Zhang, Victor Zhong, Danqi Chen, Gabor Angeli, and Christopher~D.
  Manning. 2017.
\newblock \href {https://doi.org/10.18653/v1/D17-1004} {Position-aware
  attention and supervised data improve slot filling}.
\newblock In \emph{Proceedings of the 2017 Conference on Empirical Methods in
  Natural Language Processing}, pages 35--45. Association for Computational
  Linguistics.

\bibitem[{Zhu et~al.(2015)Zhu, Kiros, Zemel, Salakhutdinov, Urtasun, Torralba,
  and Fidler}]{zhu2015aligning}
Yukun Zhu, Ryan Kiros, Rich Zemel, Ruslan Salakhutdinov, Raquel Urtasun,
  Antonio Torralba, and Sanja Fidler. 2015.
\newblock Aligning books and movies: Towards story-like visual explanations by
  watching movies and reading books.
\newblock In \emph{Proceedings of the IEEE international conference on computer
  vision}, pages 19--27.

\end{thebibliography}

\appendix

\section{Probe capacity tuning}
\label{app:hdim-tuning}
\begin{figure*}
    \centering
    \includegraphics[width=\columnwidth]{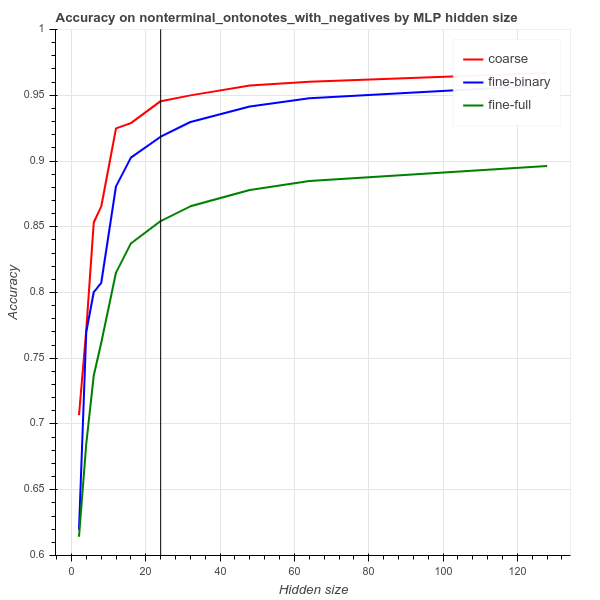}
    \includegraphics[width=\columnwidth]{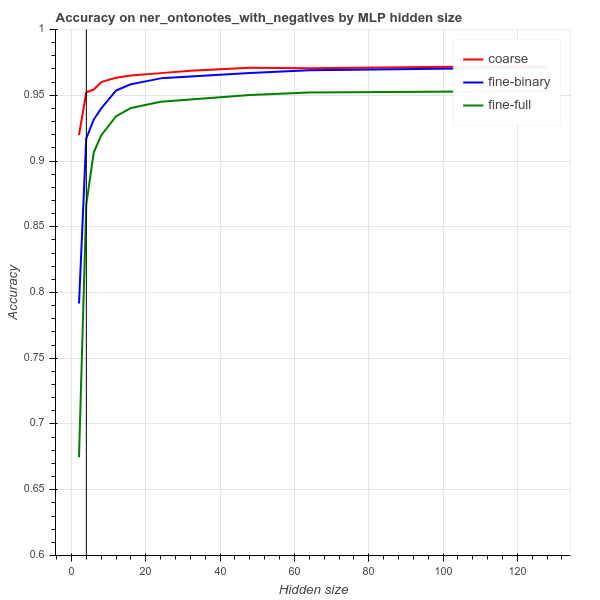}
    \includegraphics[width=\columnwidth]{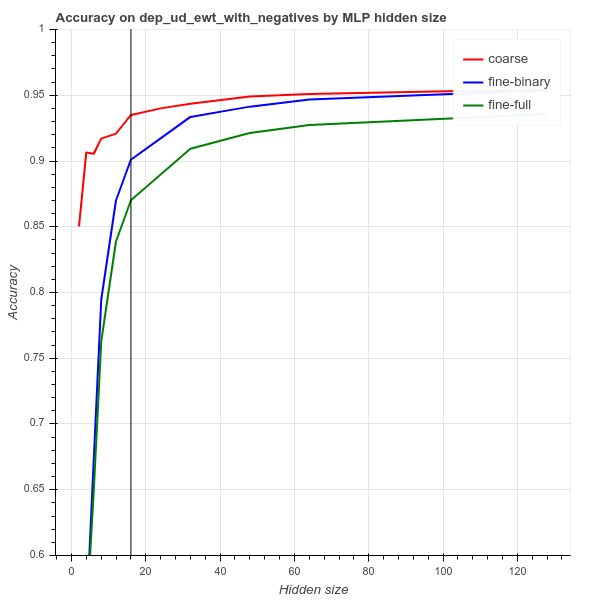}
    \includegraphics[width=\columnwidth]{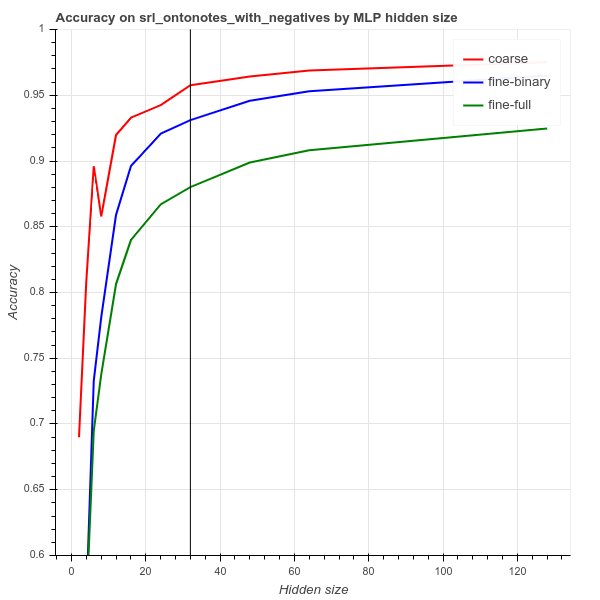}
    \caption{Performance on hidden size tuning experiments for different tasks.
    Clockwise from top-left, they are nonterminals, named entities, semantic roles, and syntactic dependencies.
    \texttt{coarse} (red) is binary accuracy of a binary classifier,
    \texttt{fine-binary} (blue) is binary accuracy of a full multiclass classifier,
    and \texttt{fine-full} (green) is the full multiclass accuracy of the multiclass classifier.
    The black vertical line is the smallest hidden size that passes the 97\% performance threshold for \texttt{coarse}.\vspace{10em}}
    \label{fig:hdim-tuning}
\end{figure*}

Results from hidden size tuning are shown in \autoref{fig:hdim-tuning}.
We use the accuracy of a binary classifier trained only on binary labels, choosing the smallest hidden size with at least 97\% of the maximum performance over all trials.
For comparison, we report the accuracy of a fully supervised multi-class model with the same hidden size.
Our method sometimes chooses a hidden size where the accuracy of the fully supervised probe is much lower than max.
While this suggests limits on the structure that can be produced, it makes our method independent of fine-grained gold labeling.
Future work may investigate the role of probe expressiveness in determining induced ontologies.

\section{More Experimental Results}
Results on larger set of encoders and tasks are shown in Tables \ref{tab:ner_ontonotes_with_negatives-loss-results}--\ref{tab:rel_tacred-loss-results}.
The extra tasks are undirected Universal Dependencies \citep{nivre2015universal},
TAC relation classification \citep{zhang2017tacred},
and OntoNotes coreference \citep{pradhan2007ontonotes}.
The extra encoders are BERT-base, multilingual BERT (mBERT)\footnote{\url{https://github.com/google-research/bert/blob/master/multilingual.md}}
and ALBERT \cite{lan2020albert}.

\section{More Analysis Results}
\begin{figure*}
  \centering
  \begin{subfigure}[b]{\textwidth}
      \centering
      \includegraphics[height=0.31\textheight]{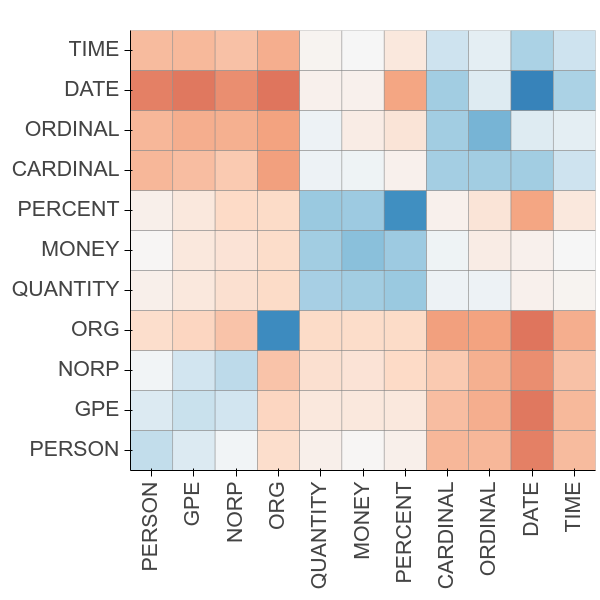}
      \includegraphics[height=0.31\textheight]{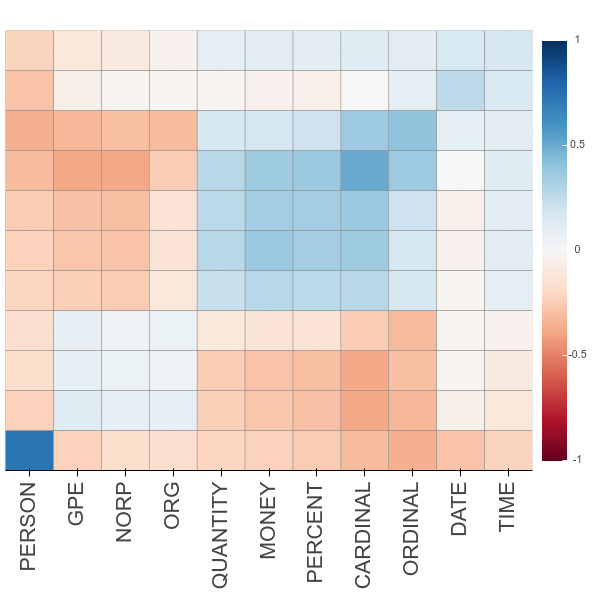}
      \caption{Pairwise \NPMI{}s for selected named entity classes in ontologies induced on BERT-large (left) and ELMo (right).}
      \label{app:fig:ner-npmi-cmp}
  \end{subfigure}
  \begin{subfigure}[b]{\textwidth}
      \centering
      \includegraphics[height=0.31\textheight]{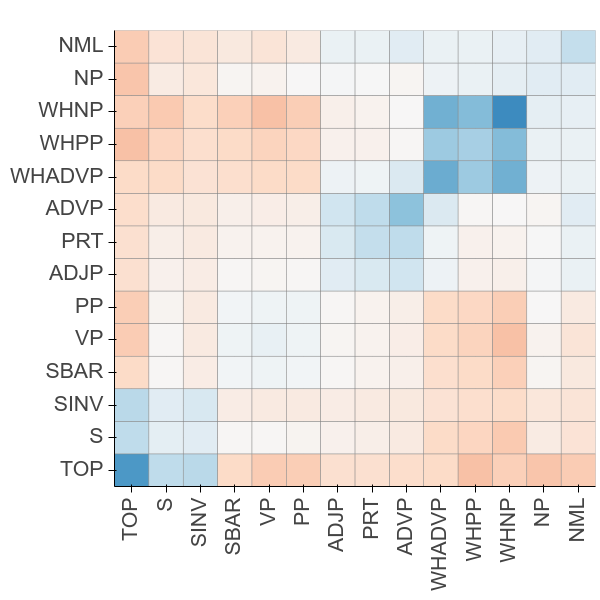}
      \includegraphics[height=0.31\textheight]{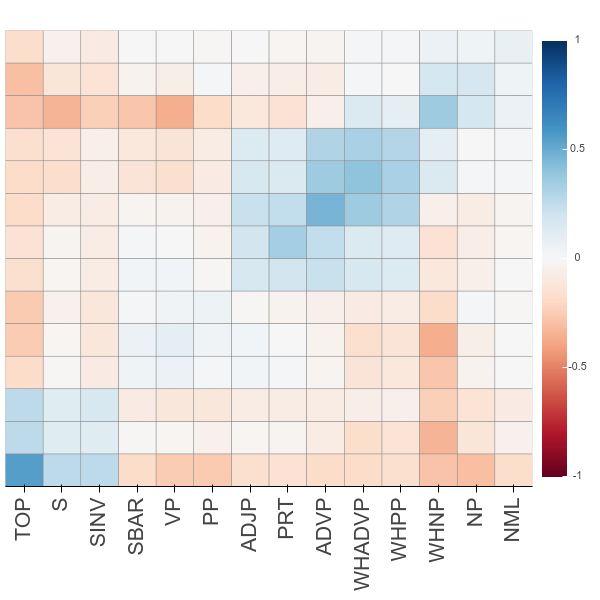}
      \caption{Pairwise \NPMI{}s for selected nonterminal classes in ontologies induced on BERT-large (left) and ELMo (right).}
      \label{app:fig:nonterminal-npmi-cmp}
  \end{subfigure}
  \caption{Pairwise \NPMI{} charts for named entities and nonterminals.}
  \label{app:fig:npmi-cmp-megafigure-1}
\end{figure*}

\begin{figure*}
  \centering
  \begin{subfigure}[b]{\textwidth}
      \centering
      \includegraphics[height=0.31\textheight]{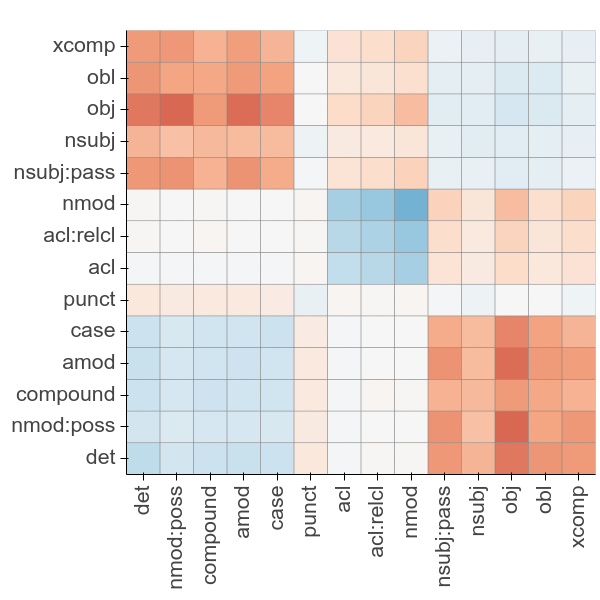}
      \includegraphics[height=0.31\textheight]{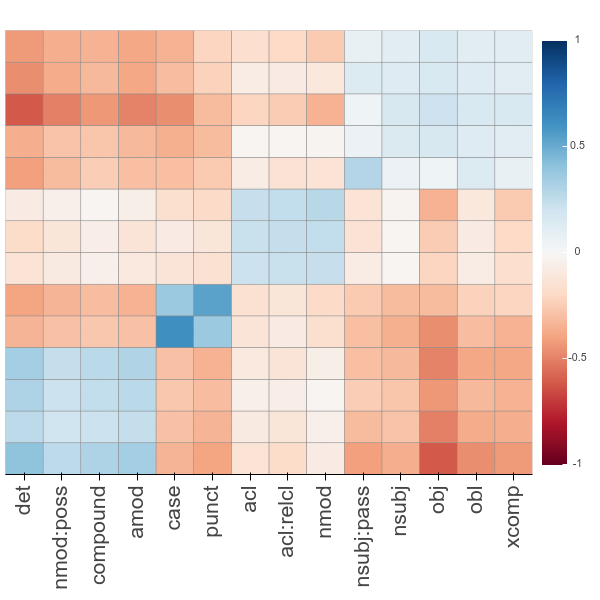}
      \caption{Pairwise \NPMI{}s for selected named universal dependency labels in ontologies induced on BERT-large (left) and ELMo (right).}
      \label{app:fig:ud-npmi-cmp}
  \end{subfigure}
  \begin{subfigure}[b]{\textwidth}
      \centering
      \includegraphics[height=0.31\textheight]{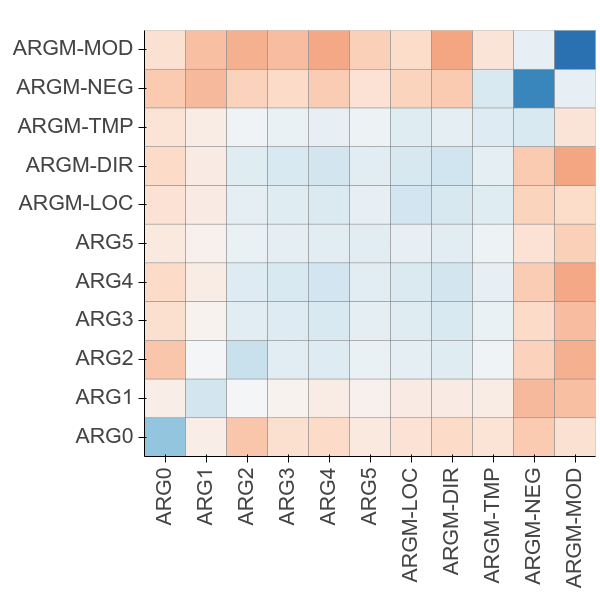}
      \includegraphics[height=0.31\textheight]{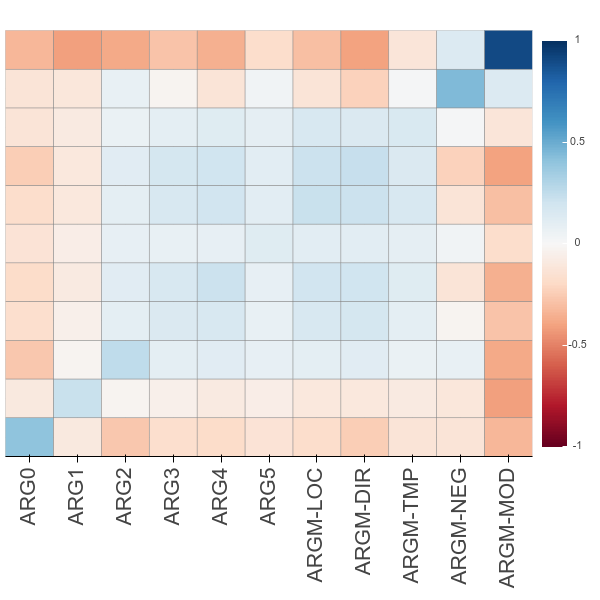}
      \caption{Pairwise \NPMI{}s for selected semantic roles in ontologies induced on BERT-large (left) and ELMo (right).}
      \label{app:fig:srl-npmi-cmp}
  \end{subfigure}
  \caption{Pairwise \NPMI{} charts for syntactic dependencies and semantic roles.}
  \label{app:fig:npmi-cmp-megafigure-2}
\end{figure*}

We show more comparative nPMI plots for BERT-large and ELMo in \autoref{app:fig:npmi-cmp-megafigure-1} and \autoref{app:fig:npmi-cmp-megafigure-2}.
These use co-occurrence counts summed over 5 runs, and exhibit the same overall trends as each run.

\begin{figure*}
  \centering
  \includegraphics[height=0.47\textheight]{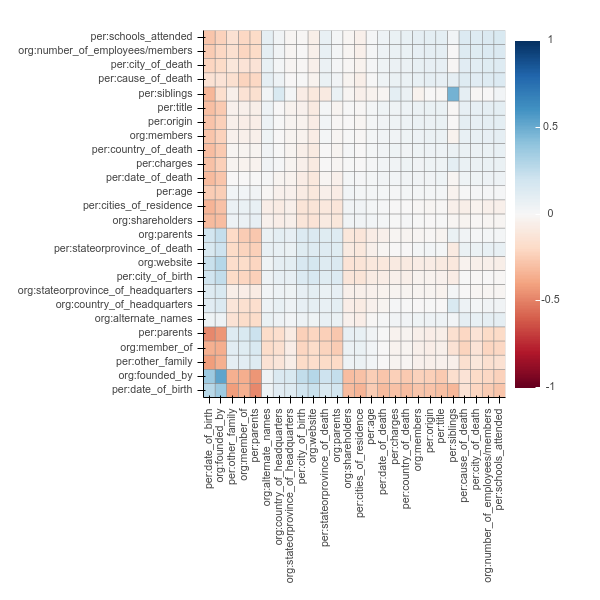}
  \includegraphics[height=0.47\textheight]{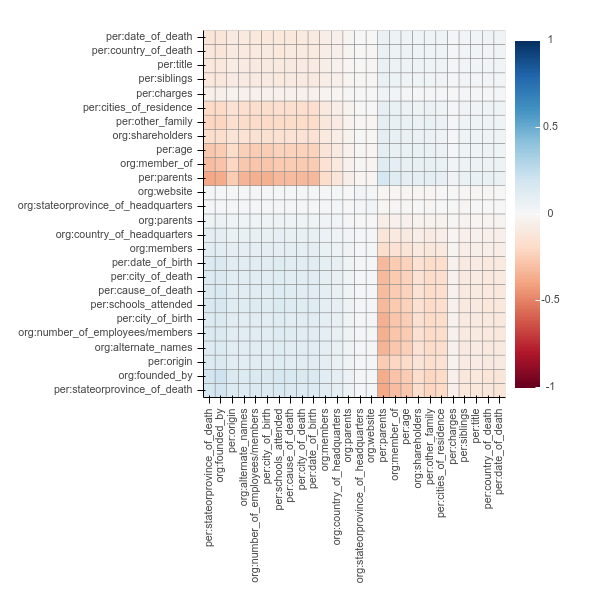}
  \caption{Pairwise \NPMI{}s for TAC relations in ontologies induced on BERT-large (top) and ELMo (bottom).}
  \label{app:fig:tacred-npmi-cmp}
\end{figure*}

\paragraph{Relation classification}
nPMI plots for BERT-large and ELMo are shown for TAC relation classification in \autoref{app:fig:tacred-npmi-cmp}. ELMo produces two diffuse groups of gold labels, while BERT seems to more clearly identify several categories of relations. Some of these may seem intuitive, \textit{e.g.}, \texttt{org:founded\_by} and \texttt{per:date\_of\_birth} relate to the creation of an entity, and are grouped together. However, the model distinguishes these from \texttt{per:origin} and \texttt{per:parents}, which may also intuitively seem similar.
The broad distribution and highly specific semantics of TAC relations makes direct qualitative assessment difficult.
Further analysis, perhaps comparing induced clusters more surface-level features (\textit{e.g.}, dependency paths) may shed more light on these results.

\begin{figure*}
  \centering
  \includegraphics[height=0.31\textheight]{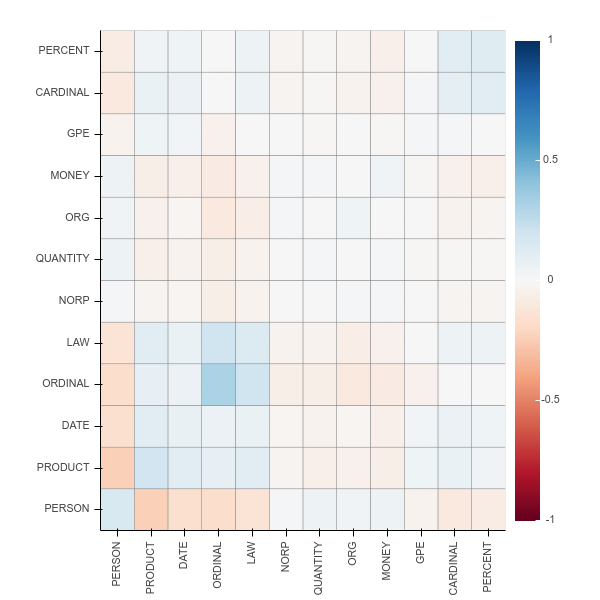}
  \includegraphics[height=0.31\textheight]{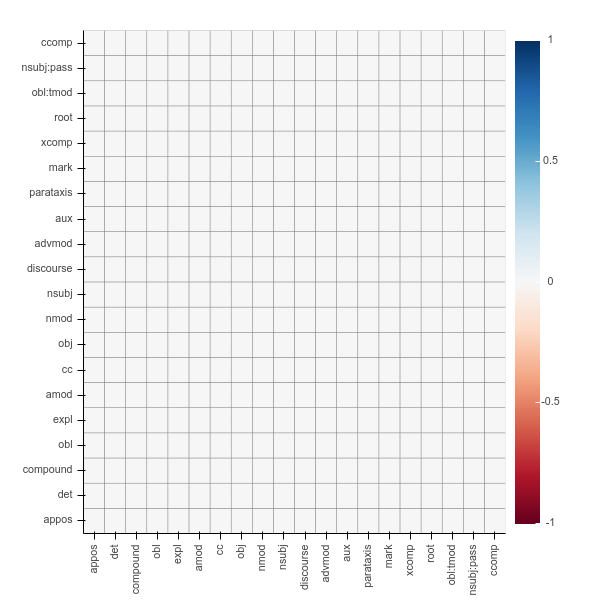}
  \includegraphics[height=0.31\textheight]{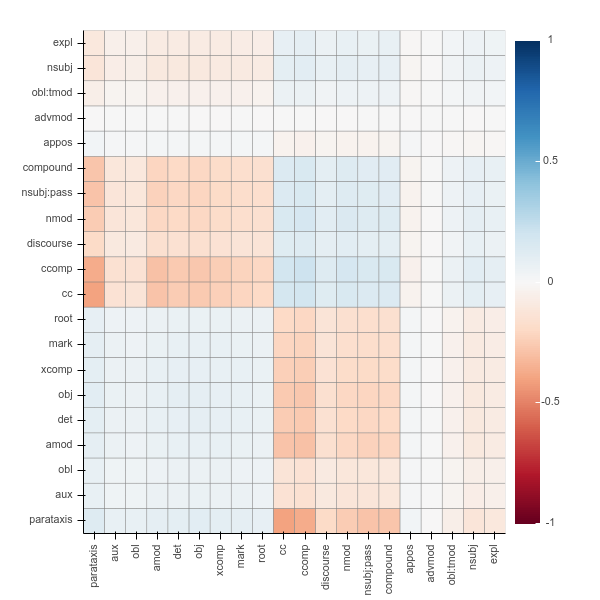}
  \includegraphics[height=0.31\textheight]{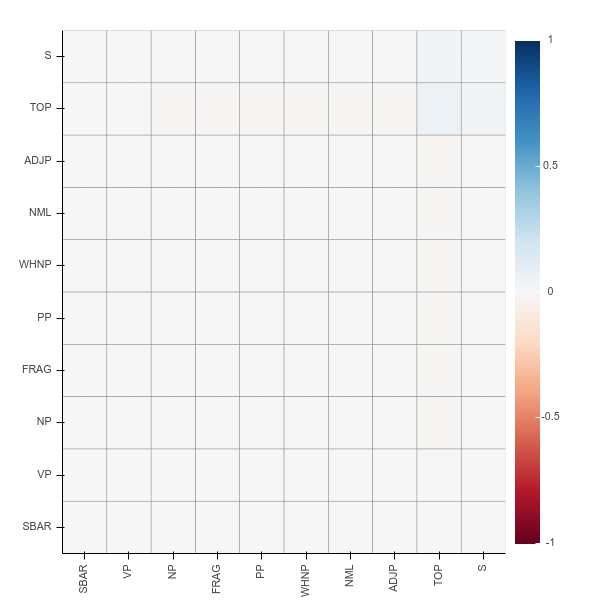}
  \includegraphics[height=0.31\textheight]{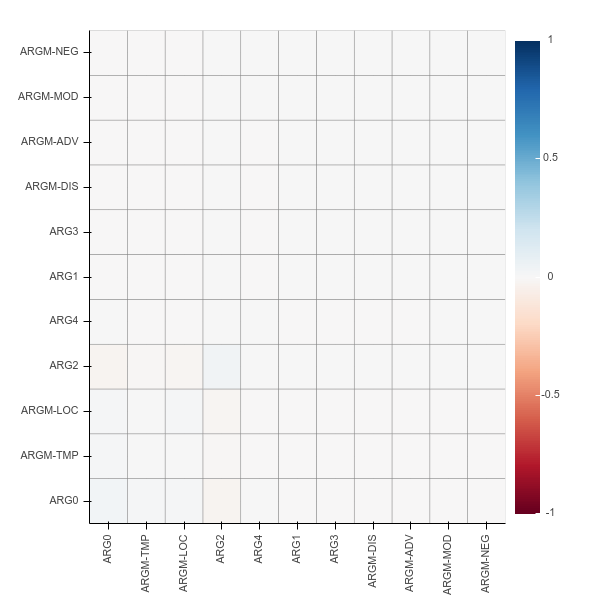}
  \includegraphics[height=0.31\textheight]{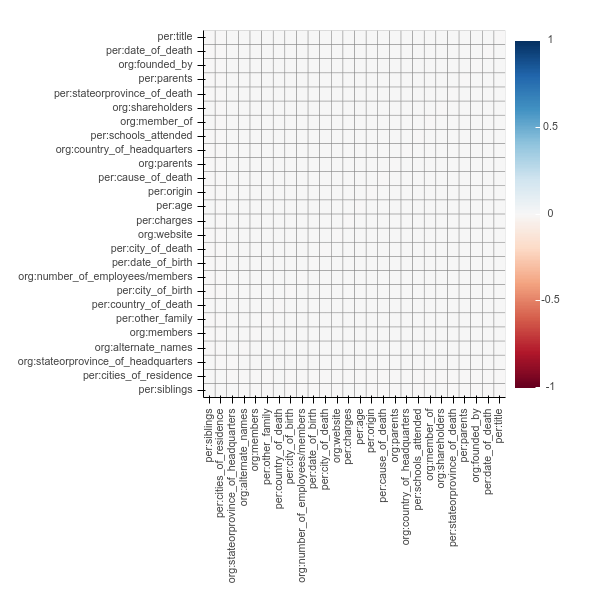}
  \caption{Pairwise \NPMI{} charts for the lexical baseline using non-contextual embeddings from BERT-large. Clockwise from top-left, they are named entities, Universal Dependencies, nonterminals, TAC relations, semantic roles, and undirected Universal Dependencies. In most cases this model seems to have no relation to gold labels, and in the few cases with interesting structure, this structure is weaker and coarser than with contextual embeddings.}
  \label{app:fig:npmi-lex-megafigure}
\end{figure*}

\paragraph{Lexical baseline results}
Normalized PMI plots for the lexical baseline on several tasks are shown in \autoref{app:fig:npmi-lex-megafigure}. 
In most cases, these show essentially no relation to gold categories. In the few cases where groups seem to emerge, they are coarser and more diffuse than what we observe with probes over contextual representations.

\begin{table*}
\centering
\begin{tabular}{lrrrrrr}
\toprule
& P & R & F1 & Acc. & \Diversity & \Uncertainty \\
\midrule
\textbf{Gold} & 1.00 & 1.00 & 1.00 & 1.00 & 9.71 & 1.00 \\
\midrule
\textbf{ELMo} & 0.40 & 0.66 & 0.50 & 0.83 & 5.07 & 1.08 \\
\textbf{BERT-base} & 0.43 & 0.57 & 0.49 & 0.88 & 6.09 & 1.11 \\
\textbf{BERT-large} & 0.47 & 0.53 & 0.50 & 0.86 & 7.50 & 1.10 \\
\textbf{mBERT} & 0.25 & 0.67 & 0.37 & 0.84 & 3.29 & 1.06 \\
\textbf{ALBERT-large} & 0.38 & 0.53 & 0.44 & 0.89 & 6.00 & 1.15 \\
\midrule
\textbf{BERT-large (lex)} & 0.19 & 0.39 & 0.26 & 0.74 & 4.33 & 1.13 \\
\bottomrule
\end{tabular}
\caption{Results by encoder for OntoNotes named entity labeling.}
\label{tab:ner_ontonotes_with_negatives-loss-results}
\end{table*}

\begin{table*}
\centering
\begin{tabular}{lrrrrrr}
\toprule
& P & R & F1 & Acc. & \Diversity & \Uncertainty \\
\midrule
\textbf{Gold} & 1.00 & 1.00 & 1.00 & 1.00 & 7.15 & 1.00 \\
\midrule
\textbf{ELMo} & 0.36 & 0.25 & 0.30 & 0.58 & 10.16 & 1.12 \\
\textbf{BERT-base} & 0.36 & 0.41 & 0.38 & 0.60 & 5.76 & 1.06 \\
\textbf{BERT-large} & 0.35 & 0.34 & 0.35 & 0.61 & 7.80 & 1.06 \\
\textbf{mBERT} & 0.36 & 0.34 & 0.35 & 0.59 & 7.38 & 1.06 \\
\textbf{ALBERT-large} & 0.38 & 0.28 & 0.32 & 0.59 & 9.07 & 1.08 \\
\midrule
\textbf{BERT-large (lex)} & 0.22 & 0.80 & 0.34 & 0.50 & 1.47 & 1.26 \\
\bottomrule
\end{tabular}
\caption{Results by encoder for OntoNotes nonterminal labeling.}
\label{tab:nonterminal_ontonotes_with_negatives-loss-results}
\end{table*}

\begin{table*}
\centering
\begin{tabular}{lrrrrrr}
\toprule
& P & R & F1 & Acc. & \Diversity & \Uncertainty \\
\midrule
\textbf{Gold} & 1.00 & 1.00 & 1.00 & 1.00 & 22.91 & 1.00 \\
\midrule
\textbf{ELMo} & 0.23 & 0.42 & 0.29 & 0.67 & 11.11 & 1.22 \\
\textbf{BERT-base} & 0.13 & 0.34 & 0.19 & 0.76 & 9.69 & 1.23 \\
\textbf{BERT-large} & 0.14 & 0.33 & 0.19 & 0.77 & 11.22 & 1.23 \\
\textbf{mBERT} & 0.27 & 0.51 & 0.35 & 0.73 & 9.40 & 1.22 \\
\textbf{ALBERT-large} & 0.23 & 0.41 & 0.29 & 0.72 & 9.84 & 1.20 \\
\midrule
\textbf{BERT-large (lex)} & 0.06 & 0.86 & 0.11 & 0.50 & 1.33 & 1.02 \\
\bottomrule
\end{tabular}
\caption{Results by encoder for Universal Dependency labeling.}
\label{tab:dep_ud_ewt_with_negatives-loss-results}
\end{table*}

\begin{table*}
\centering
\begin{tabular}{lrrrrrr}
\toprule
& P & R & F1 & Acc. & \Diversity & \Uncertainty \\
\midrule
\textbf{Gold} & 1.00 & 1.00 & 1.00 & 1.00 & 22.91 & 1.00 \\
\midrule
\textbf{ELMo} & 0.19 & 0.23 & 0.21 & 0.71 & 19.12 & 1.14 \\
\textbf{BERT-base} & 0.27 & 0.24 & 0.25 & 0.85 & 22.79 & 1.20 \\
\textbf{BERT-large} & 0.23 & 0.23 & 0.23 & 0.82 & 18.51 & 1.17 \\
\textbf{mBERT} & 0.24 & 0.20 & 0.21 & 0.83 & 20.31 & 1.19 \\
\textbf{ALBERT-large} & 0.30 & 0.27 & 0.28 & 0.81 & 20.53 & 1.14 \\
\midrule
\textbf{BERT-large (lex)} & 0.09 & 0.54 & 0.16 & 0.50 & 3.39 & 1.00 \\
\bottomrule
\end{tabular}
\caption{Results by encoder for undirected Universal Dependency labeling.}
\label{tab:dep_ud_ewt_undirected_with_negatives-loss-results}
\end{table*}

\begin{table*}
\centering
\begin{tabular}{lrrrrrr}
\toprule
& P & R & F1 & Acc. & \Diversity & \Uncertainty \\
\midrule
\textbf{Gold} & 1.00 & 1.00 & 1.00 & 1.00 & 8.73 & 1.00 \\
\midrule
\textbf{ELMo} & 0.40 & 0.17 & 0.24 & 0.76 & 22.35 & 1.08 \\
\textbf{BERT-base} & 0.39 & 0.18 & 0.25 & 0.86 & 21.95 & 1.15 \\
\textbf{BERT-large} & 0.37 & 0.17 & 0.24 & 0.88 & 18.70 & 1.15 \\
\textbf{mBERT} & 0.41 & 0.21 & 0.28 & 0.88 & 19.05 & 1.12 \\
\textbf{ALBERT-large} & 0.43 & 0.21 & 0.28 & 0.87 & 19.90 & 1.12 \\
\midrule
\textbf{BERT-large (lex)} & 0.19 & 0.39 & 0.26 & 0.46 & 2.81 & 1.01 \\
\bottomrule
\end{tabular}
\caption{Results by encoder for OntoNotes semantic role labeling.}
\label{tab:srl_ontonotes_with_negatives-loss-results}
\end{table*}

\begin{table*}
\centering
\begin{tabular}{lrrrrrr}
\toprule
& P & R & F1 & Acc. & \Diversity & \Uncertainty \\
\midrule
\textbf{Gold} & 1.00 & 1.00 & 1.00 & 1.00 & 1.00 & 1.00 \\
\midrule
\textbf{ELMo} & 1.00 & 0.09 & 0.16 & 0.80 & 14.22 & 1.18 \\
\textbf{BERT-base} & 1.00 & 0.09 & 0.16 & 0.86 & 14.67 & 1.24 \\
\textbf{BERT-large} & 1.00 & 0.09 & 0.17 & 0.87 & 15.57 & 1.27 \\
\textbf{mBERT} & 1.00 & 0.09 & 0.16 & 0.83 & 13.86 & 1.24 \\
\textbf{ALBERT-large} & 1.00 & 0.09 & 0.16 & 0.86 & 13.56 & 1.26 \\
\midrule
\textbf{BERT-large (lex)} & 1.00 & 0.78 & 0.87 & 0.78 & 1.60 & 1.03 \\
\bottomrule
\end{tabular}
\caption{Results by encoder for OntoNotes coreference. Note the high diversity scores, showing that the LSL model can find fine-grained structure even in the case of binary labels.}
\label{tab:coref_unified_ontonotes-loss-results}
\end{table*}

\begin{table*}
\centering
\begin{tabular}{lrrrrrr}
\toprule
& P & R & F1 & Acc. & \Diversity & \Uncertainty \\
\midrule
\textbf{Gold} & 1.00 & 1.00 & 1.00 & 1.00 & 24.78 & 1.00 \\
\midrule
\textbf{ELMo} & 0.11 & 0.78 & 0.20 & 0.77 & 2.38 & 1.05 \\
\textbf{BERT-base} & 0.11 & 0.90 & 0.20 & 0.76 & 1.94 & 1.05 \\
\textbf{BERT-large} & 0.16 & 0.63 & 0.25 & 0.80 & 3.87 & 1.11 \\
\textbf{mBERT} & 0.15 & 0.87 & 0.26 & 0.76 & 2.21 & 1.05 \\
\midrule
\textbf{BERT-large (lex)} & 0.07 & 0.97 & 0.13 & 0.76 & 1.11 & 1.02 \\
\bottomrule
\end{tabular}
\caption{Results by encoder for TAC relation classification. Note that the diversity scores are much lower than gold for most encoders. This accords with \citet{tenney2019what}'s findings that ELMo and BERT have middling performance on the task; it seems unlikely that the highly specific relations in TACRED are salient in their feature spaces.}
\label{tab:rel_tacred-loss-results}
\end{table*}

\end{document}